\documentclass[preprint,12pt,authoryear]{elsarticle}

\usepackage{amssymb}
\usepackage[utf8]{inputenc}
\usepackage{color,xcolor}
\usepackage{setspace}
\usepackage[margin=1.25in]{geometry}
\usepackage{graphicx}
\usepackage{subcaption}
\usepackage{amsmath}
\usepackage{lineno}
\usepackage{url}
\usepackage{makecell}
\usepackage{multirow}
\usepackage{lscape}
\usepackage{longtable}
\usepackage{arydshln}
\usepackage{soul}
\usepackage{enumitem}
\usepackage{textgreek}
\usepackage[acronym, automake]{glossaries}

\journal{ }

\makeglossaries

\newacronym{ml}{ML}{Machine Learning} 
\newacronym{swir}{SWIR}{Short-Wave Infrared} 
\newacronym{cnn}{CNN}{Convolutional Neural Networks} 
\newacronym{tir}{TIR}{Thermal Infrared} 
\newacronym{sbmp}{SBMP}{Single-Band-Multiple-Pass} 
\newacronym{mbsp}{MBSP}{Multi-Band-Single-Pass}
\newacronym{mbmp}{MBMP}{Multi-Band-Multi-Pass}
\newacronym{ime}{IME}{Integrated Mass Enhancement}
\newacronym{gsd}{GSD}{Ground Sampling Distance}
\newacronym{iss}{ISS}{International Space Station} 
\newacronym{emit}{EMIT}{Earth Surface Mineral Dust Source Investigation}
\newacronym{aviris}{AVIRIS}{Airborne Visible Infrared Imaging Spectrometer} 
\newacronym{avirisng}{AVIRIS-NG}{Airborne Visible Infrared Imaging Spectrometer - Next Generation}
\newacronym{gmi}{GMI}{Greenhouse-gases Monitoring Instrument} 
\newacronym{ahsi}{AHSI}{Advanced Hyperspectral Imager} 
\newacronym{goes}{GOES}{Geostationary Operational Environmental Satellites} 
\newacronym{gosat}{GOSAT}{Greenhouse Gases Observing Satellite} 
\newacronym{mtg}{MTG}{Meteosat Third Generation} 
\newacronym{cnsa}{CNSA}{China National Space Agency} 
\newacronym{cresda}{CRESDA}{China Center for Resources Satellite Data and Application} 
\newacronym{cast}{CAST}{China Academy of Space Technology} 
\newacronym{esa}{ESA}{European Space Agency} 
\newacronym{noaa}{NOAA}{National Oceanic and Atmospheric Administration} 
\newacronym{nasa}{NASA}{National Aeronautics and Space Administration} 
\newacronym{jpl}{JPL}{Jet Propulsion Laboratory} 
\newacronym{jaxa}{JAXA}{Japan Aerospace Exploration Agency} 
\newacronym{nzsa}{NZSA}{New Zealand Space Agency} 
\newacronym{edf}{EDF}{European Defence Fund}
\newacronym{asi}{ASI}{Italian Space Agency}
\newacronym{dlr}{DLR}{German Aerospace Center}
\newacronym{usgs}{USGS}{United States Geological Survey}
\newacronym{eol}{EOL}{End of Life}
\newacronym{fwhm}{FWHM}{Full Width at Half Maximum}
\newacronym{gf}{GF}{GaoFen}
\newacronym{hj}{HJ}{Huanjing}
\newacronym{s5p}{Sentinel 5P}{Sentinel-5 Precursor}
\newacronym{abi}{ABI}{Advanced Baseline Imager}
\newacronym{tanso-fts}{TANSO-FTS}{Thermal And Near infrared Sensor for carbon Observations - Fourier Transform Spectrometer }
\newacronym{viirs}{VIIRS}{Visible Infrared Imaging Radiometer Suite}
\newacronym{waf-p}{WAF-P}{Wide-Angle Fabry-Perot}
\newacronym{prisma}{PRISMA}{Pecursore Iperspettrale della Missione Applicativa}
\newacronym{hyp}{HYP}{Hyperspectral}
\newacronym{enmap}{EnMAP}{Environmental Mapping and Analysis Program}
\newacronym{oli}{OLI}{Operational Land Imager}
\newacronym{msi}{MSI}{Multispectral Instrument}
\newacronym{geisatp}{GeiSAT P}{GeiSAT Precursor}
\newacronym{isim}{iSim}{Integrated Standard Imager for Microsatellites}
\newacronym{ghost}{GHOSt}{Global Hyperspectral Observation Satellite}
\newacronym{hi}{HI}{Hyperspectral imager}
\newacronym{ch4}{CH\textsubscript{4}}{Chemical formula for Methane}
\newacronym{co2}{CO\textsubscript{2}}{Chemical formula for Carbon Dioxide}
\newacronym{merlin}{MERLIN}{Methane Remote Sensing Lidar Mission}
\newacronym{gosatgw}{GOSAT-GW}{Global Observing Satellite for Greenhouse gases and Water cycle}
\newacronym{tanso3}{TANSO-3}{Total Anthropogenic and Natural emissions mapping SpectrOmeter-3}
\newacronym{tango}{Tango}{Twin Anthropogenic Greenhouse Gas Observers}
\newacronym{co2m}{CO2M}{Copernicus Anthropogenic Carbon Dioxide Monitoring constellation}
\newacronym{chime}{CHIME}{Copernicus Hyperspectral Imaging Mission}
\newacronym{xchco4}{XCHCO\textsubscript{4}}{xxxxxx}
\newacronym{xch4}{XCH\textsubscript{4}}{column average dry air mixing ratio of methane}
\newacronym{modtrans}{MODTRAN}{Moderate resolution atmospheric Transmission}
\newacronym{csf}{CSF}{Cross-Sectional Flux}
\newacronym{rs}{RS}{Remote Sensing}
\newacronym{sron}{SRON}{Netherlands Institute for Space Research}
\newacronym{ecmwf}{ECMWF}{European Centre for Medium-Range Weather Forecasts}
\newacronym{gfs}{GFS}{Global Forecast System}
\newacronym{nws}{NWS}{American National Weather Service}
\newacronym{imeo}{IMEO}{International Methane Emissions Observatory}
\newacronym{mars}{MARS}{Methane Alert and Response System}
\newacronym{edgarv6}{EDGAR V6}{Emissions Database for Global Atmospheric Research Version 6}
\newacronym{gfei}{GFEI}{Global Fuel Emissions Inventory}
\newacronym{mlp}{MLP}{Multi-Layer Perceptrons}
\newacronym{svm}{SVM}{Support Vector Machines}
\newacronym{vgg-16}{VGG-16}{Visual Geometry Group 16}
\newacronym{nox}{NO\textsubscript{X}}{Nitrogen oxides}
\newacronym{snr}{SNR}{Signal-to-Noise Ratio}
\newacronym{rgb}{RGB}{Red-green-blue}
\newacronym{coco}{COCO}{Common Objects in Context}
\newacronym{detr}{DETR}{Detection transformer}
\newacronym{gao}{GAO}{Global Airborne Observatory}
\newacronym{iou}{IoU}{Intersection over Union}
\newacronym{fpr}{FPR}{False Positive Rate}
\newacronym{miou}{mIoU}{mean Intersection over Union}
\newacronym{map}{mAP}{median Average Precision}
\newacronym{ap}{AP}{Average Precision}
\newacronym{fnr}{FNR}{False Negative Rate}
\newacronym{auprc}{AUPRC}{Area under the Precision-Recall Curve}
\newacronym{tpr}{TPR}{True Positive Rate}
\newacronym{rmse}{RMSE}{Root mean squared Error}
\newacronym{are}{ARE}{Absolute Relative Error}
\newacronym{mlr}{MLR}{Multiple Linear Regression}
\newacronym{mape}{MAPE}{Mean Absolute Percentage Error}
\newacronym{mae}{MAE}{Mean Absolute Error}
\newacronym{r}{R}{Pearson correlation coefficient}
\newacronym{posp}{POSP}{Particulate Observing Scanning Polarimeter}
\newacronym{tropomi}{TROPOMI}{TROPOspheric Monitoring Instrument}
\newacronym{slstr}{SLSTR}{Sea and Land Surface Temperature Radiometer}
\newacronym{hyc}{HYC}{HYperspectral Camera}
\newacronym{hsi}{HSI}{Hyperspectral Imager}
\newacronym{modtran}{MODTRAN}{Moderate resolution atmospheric Transmission}

\begin{document}

\begin{frontmatter}

\title{Machine Learning for Methane Detection and Quantification from Space - A survey}

\author[a,b,*]{Enno Tiemann}
\author[c]{Shanyu Zhou}
\author[b]{Alexander Kläser}
\author[a]{Konrad Heidler} 
\author[d]{Rochelle Schneider}
\author[a,e]{Xiao Xiang Zhu}

\affiliation[a]{organization={Data Science in Earth Observation, Technical University of Munich},
           addressline={Arcisstraße 21}, 
           city={Munich},
           postcode={80333}, 
           country={Germany}}
           
\affiliation[b]{organization={OHB Digital Connect GmbH},
           city={Bremen}, 
           country={Germany}}
           
\affiliation[c]{organization={University of Valencia},
           city={Valencia}, 
           country={Spain}}

\affiliation[d]{organization = {$\Phi$-lab, European Space Agency},
           city={Frascati},
           country={Italy}}

\affiliation[e]{organization = {Munich Center for Machine Learning},
           city={Munich},
           country={Germany}}
           
\affiliation[*]{organization={Address correspondence to: enno.tiemann@tum.de}}

\begin{abstract}

Methane (\acrshort{ch4}) is a potent anthropogenic greenhouse gas, contributing 86 times more to global warming than Carbon Dioxide (\acrshort{co2}) over 20 years, and it also acts as an air pollutant. Given its high radiative forcing potential and relatively short atmospheric lifetime (9±1 years), methane has important implications for climate change, therefore, cutting methane emissions is crucial for effective climate change mitigation.

This work expands existing information on operational methane point source detection sensors in the Short-Wave Infrared (SWIR) bands. It reviews the state-of-the-art for traditional as well as Machine Learning (ML) approaches. The architecture and data used in such ML models will be discussed separately for methane plume segmentation and emission rate estimation. 
Traditionally, experts rely on labor-intensive manually adjusted methods for methane detection. However, ML approaches offer greater scalability. Our analysis reveals that ML models outperform traditional methods, particularly those based on convolutional neural networks (CNN), which are based on the U-net and transformer architectures. These ML models extract valuable information from methane-sensitive spectral data, enabling a more accurate detection. 
Challenges arise when comparing these methods due to variations in data, sensor specifications, and evaluation metrics. To address this, we discuss existing datasets and metrics, providing an overview of available resources and identifying open research problems. 
Finally, we explore potential future advances in ML, emphasizing approaches for model comparability, large dataset creation, and the European Union’s forthcoming methane strategy. 

\end{abstract}

\end{frontmatter}


\section{Introduction}

Methane (\acrshort{ch4}) is the second most significant anthropogenic greenhouse gas and a precursor to tropospheric ozone, which functions as both another greenhouse gas and an air pollutant \citep{comission_workfile_2021, rouet2024automatic}. Methane is also a potential air pollutant, contributing to premature deaths and reductions in agricultural yields \citep{staniaszek2022role, comission_workfile_2021}. Due to its high radiative forcing potential and relatively short atmospheric lifetime of approximately 9±1 years, mitigating methane emissions is prioritized as an effective strategy for combating climate change over decades \citep{etminan2016radiative, irakulis2021satellite, prather2012reactive, eyeonmethane_IMEO2022, shindell2012simultaneously}. Methane has accounted for about 30\% of observed global warming \citep{IEAglobalmethanetracker2022}, with a global warming potential 86 times that of \acrshort{co2} over 20 years \citep{comission_workfile_2021}. Atmospheric methane levels have nearly tripled since preindustrial times \citep{irakulis2021satellite}. 
The most significant sources of anthropogenic methane emissions are agriculture (approximately 142 Mt), the energy sector (oil, gas, coal, and bioenergy) with 128 Mt, and the waste sector with 71 Mt as of 2023 \citep{IEAglobalmethanetracker2024}. Emissions exceeding 25 kg/h, known as super-emitters, provide a substantial impact on the overall methane budget \citep{vaughan2024ch4net, zavala2015toward}. In the US natural gas supply chain, super-emitters account for over 60\% of the energy sector's total emissions \citep{duren2019california}. Thus, reducing super-emitters is crucial for rapid, cost-effective climate change mitigation \citep{lauvaux2022global}. Consequently, over 110 countries have signed the Global Methane Pledge, committing to a 30\% reduction in methane emissions by 2030 \citep{jacob2022quantifying}. The European Union has supported this initiative by funding the International Methane Emissions Observatory (IMEO) and the Methane Alert and Response System (MARS), underscoring the current emphasis on methane monitoring. 
Methane emissions in the oil and gas sector result from extraction, processing, and delivery through both intentional (e.g., venting and flaring) and unintentional (e.g., equipment leaks) activities \citep{eyeonmethane_IMEO2022}. Emissions are typically calculated using bottom-up approaches, starting from the average emissions of facilities multiplied by the number of such facilities \citep{jacob2022quantifying}. Various inventories, such as the Global Fuel Emissions Inventory (GFEI) \citep{scarpelli2022updated} and Emission Database for Global Atmospheric Research (EDGAR) V6 \citep{EDGARV6_dataset}, are available for different sectors. Still, these inventories often underestimate emissions compared to top-down global observations, leading to significant uncertainties in policy decision-making \citep{thorpe2020methane, chen2022quantifying, yu2022methane, rouet2024automatic}. 

Satellites are frequently used as fast, cost-effective detection systems for top-down methane inventories due to their high spatial coverage and ability to observe large areas and individual super-emitter plumes. 
Two types of satellite instruments are employed for methane detection and quantification: infrared sounders in the \acrfull{tir} range (8000-15000 nm) for vertical methane profiles and spectrometers in the \acrfull{swir} range (1400-3000 nm) for surface emissions \citep{worden2013ch, gorrono2023understanding, pandey2023daily, zhang2022detecting, irakulis2022satellites}. \acrshort{tir} instruments are generally used for methane estimation in the upper and middle atmosphere, while \acrshort{swir} instruments are more suitable for detecting surface emissions \citep{li2024high}. 
This study focuses on surface emissions, specifically, those spectroscopy systems operating in the \acrshort{swir} range. Within the \acrshort{swir} range, there are two significant absorption bands at approximately 1650 nm and 2300 nm, necessitating different approaches for detection due to varying gas absorptions \citep{li2024high}. 

Methane detection can be divided into three tasks: (1) detecting and calculating methane column concentration per pixel, (2) segmenting methane plumes, and (3) estimating the emission rate and location of the emitting source \citep{jacob2016satellite, jacob2022quantifying}. Traditional methods and \acrfull{ml} approaches differ significantly in tackling these tasks. 

Traditional methods for detecting and calculating methane column concentration per pixel involve the physical inversion of measured radiance using techniques such as band ratios \citep{ruuvzivcka2023semantic}, multi-band multi-pass methods \citep{varon2021high}, and linear multiple regression for multispectral instruments \citep{sanchez2022mapping, ruuvzivcka2023semantic}. Hyperspectral instruments often employ matched filter methods \citep{foote2020fast, irakulis2021satellite, guanter2021mapping, thorpe2016mapping}. These approaches are typically computationally expensive and require manual inspection. 
In contrast, \acrshort{ml} methods provide scalability for the detection process, utilizing data from multispectral radiance \citep{rouet2024automatic}, methane column concentrations \citep{radman2023s2metnet, vaughan2024ch4net, ruuvzivcka2023semantic}, band ratios \citep{rouet2023autonomous}, processed hyperspectral data \citep{ruuvzivcka2023semantic, jongaramrungruang2022methanet}, and hyperspectral radiance data \citep{kumar2023methanemapper, kumar2020deep, joyce2023using}. 

Traditional methods for segmenting methane plumes rely on predefined rules and thresholds. These methods use concentration thresholds, wind direction, minimum plume size, and other manually designed features to identify and segment plumes \citep{guanter2021mapping, jacob2022quantifying, gorrono2023understanding}. 
\acrshort{ml} approaches, on the other hand, use advanced algorithms to automatically segment methane plumes if learned from training data. These methods can handle multispectral radiance data, methane column concentrations, and other relevant data types without the need for manual threshold setting, making the segmentation process more efficient and accurate \citep{schuit2023automated, vanselow2024automated, jongaramrungruang2022methanet, rouet2024automatic}.

Estimating emission rates traditionally employs the \acrfull{ime} method, which calculates emission rates based on methane column concentration, plume size, and wind information \citep{frankenberg2016airborne, varon2018quantifying}. This method can be accurate but is often labor-intensive and requires the availability of high-quality input data. 
\acrshort{ml} methods streamline this process by estimating emission rates directly from enhancement maps or using inputs similar to those in the \acrshort{ime} model. These approaches can quickly analyze large datasets and provide accurate emission rate estimates without extensive manual involvement and checks \citep{jongaramrungruang2022methanet, radman2023s2metnet, bruno2023u}.

The current state-of-the-art in the literature suggests that \acrshort{ml} approaches offer scalable, automated, efficient, and potentially more accurate alternatives, utilizing various types of radiance and concentration data to achieve their tasks while traditional methods for methane detection, segmentation, and emission rate estimation rely heavily on manual processes and predefined rules.

While \acrshort{ml} models rely on different architectures, a common challenge persists in the availability of large amounts of high-quality training data. Researchers have adopted different strategies to provide training data in the context of plume mask generation. Some authors rely on manually annotated masks \citep{schuit2023automated, vaughan2024ch4net, kumar2023methanemapper, ruuvzivcka2023semantic}, while others utilize simulated data for training \citep{bruno2023u, jongaramrungruang2022methanet, radman2023s2metnet, joyce2023using, rouet2024automatic}. However, when it comes to estimating emission rates, the situation becomes more challenging. Ground truth data for emission rates is scarce, as it is primarily derived from controlled release experiments \citep{sherwin2024single}. Unfortunately, this limited ground truth data is inadequate for \acrshort{ml} training purposes. Consequently, researchers have turned to simulated datasets, which suffer from sparsity regarding sample diversity. They lack variation in backgrounds, geographical locations, plume characteristics, wind speeds, emission rates, and real-world data. Addressing these limitations is crucial for advancing \acrshort{ml}-based plume analysis and emission estimation.
 
Most \acrshort{ml} models for methane tasks are task-specific (e.g. only for plume segmentation), allowing the use of traditional methods for other tasks (e.g. emission rate estimation), often involving manual expert verification of the \acrshort{ml} output. These models, primarily based on \acrfull{cnn}, have demonstrated enhanced detection capabilities and accuracy compared to traditional methods \citep{schuit2023automated, bruno2023u, jongaramrungruang2022methanet, rouet2024automatic, vaughan2024ch4net, ruuvzivcka2023semantic, joyce2023using}. However, as mentioned above, they require large amounts of training data and are specialized to specific sensors. 

This work focuses on \acrshort{ml} approaches for methane detection and extends previous studies \citep{jacob2016satellite, jacob2022quantifying} by reviewing state-of-the-art methods and recent developments. In addition to the previous studies, we expand the current and future sensors for methane monitoring and provide a comparison for methane detection. 
Next, this review provides a first comprehensive review of \acrshort{ml} methods, and it reviews used metrics for comparability of different approaches and available datasets in this context. In the end, this work compares traditional methods and discusses future trends and approaches for improving \acrshort{ml} models.

\section{Sensors for Methane Monitoring}
\label{chap:satellite_systems}
\begin{figure}[h]
    \centering
    \includegraphics[width=1\textwidth]{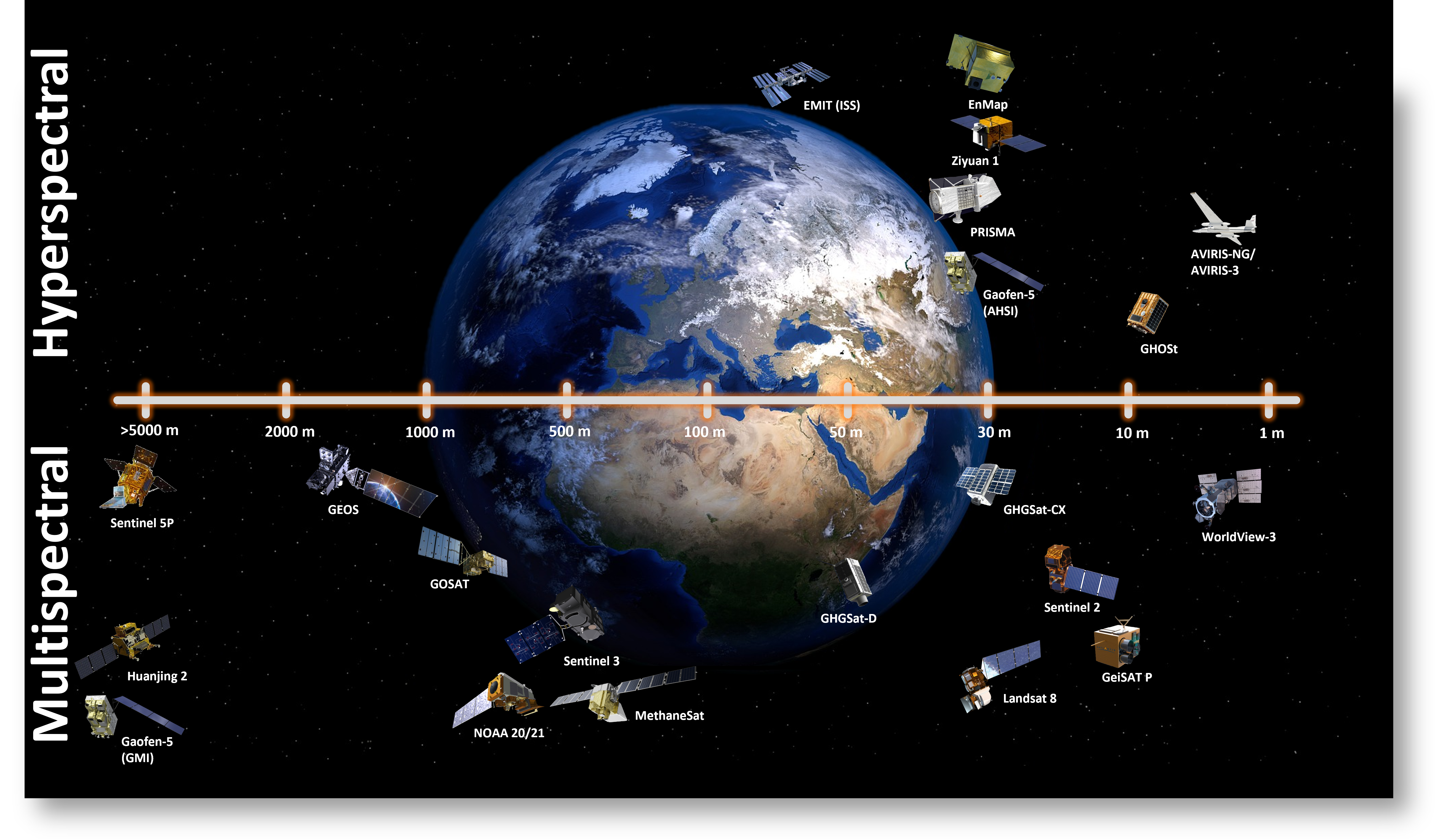}
    \caption{Overview of existing methane monitoring systems with categorization according to \acrshort{gsd} and categorization into multi- and hyperspectral missions.}
    \label{fig:obs_sys_overview}
\end{figure}

This section summarizes and discusses the operational satellites and instruments utilized for methane monitoring. The systems can be broadly classified into \textit{area flux mappers} and \textit{point-source mappers}.
 \textit{Area flux mappers} observe large areas to detect changes at country or regional levels, focusing on measuring more significant methane emissions. Satellites such as \acrfull{s5p}, \acrfull{gosat}, and Sentinel 3 fall into this category.
 On the other hand, \textit{point-source mappers} are designed to observe single plumes and estimate their size and emission rate. These instruments offer smaller observation areas but higher spatial resolution, enabling the detection of smaller plumes. Satellites like Sentinel 2, Landsat 8/9, Worldview-3, GHG-Sat constellation, \acrfull{prisma}, \acrfull{enmap}, and \acrfull{avirisng} are examples of \textit{point-source mappers}.
 Furthermore, methane monitoring systems can be categorized based on the number of spectral bands they employ. Multispectral systems utilize a few larger bands, while hyperspectral systems utilize many bands with small bandwidths. Examples of multispectral systems include \acrshort{s5p}, \acrfull{goes}, \acrshort{gosat}, Sentinel 3, Sentinel 2, Landsat 8 and 9, Worldview-3, and GHGSat. Hyperspectral systems encompass \acrshort{prisma}, \acrshort{enmap}, \acrshort{avirisng}, \acrfull{emit}, and others in development.
 For multispectral satellites, spatial resolution is crucial for detecting plume sizes, while area flux mappers prioritize spectral resolution for accurate measurements at the cost of spatial resolution.
 The different missions are illustrated in Figure \ref{fig:obs_sys_overview} and plotted by \acrfull{gsd} with categorization into hyper-/multispectral.

Figure \ref{fig:obs_sys_overview} shows that satellites designed for larger areas are predominantly multispectral, offering limited bands tailored to each specific product and use case. Another finding is that hyperspectral satellites primarily feature a 30 m spatial resolution (with exceptions such as \acrshort{emit} mounted on the \acrfull{iss}, Aurora as a commercial satellite and \acrshort{avirisng} as an airborne sensor). This illustrates the trade-off between spatial and spectral resolution, aiming to strike a balance that ensures adequate resolution for widely used applications in remote sensing. The Chinese GaoFen 5 satellite is listed two times in Figure \ref{fig:obs_sys_overview} as the satellite carries two instruments for methane observation (the multispectral \acrfull{gmi} and the hyperspectral \acrfull{ahsi} instruments).

Detailed information on the different systems for methane observations is listed in Table \ref{tab:all_sensors}. All mentioned systems measure the radiance of solar backscatter in the \acrshort{swir} band for methane detection, with standard spectral bands ranging from 1630 to 1700 nm (1650 nm band) or 2200 to 2400 nm (2300 nm band) \citep{jacob2022quantifying}. All satellites operate on a sun-synchronous, low-earth orbit except for \acrshort{goes}, which is on a geostationary orbit, providing images only during daylight hours. 
Operational modes typically involve either pointing capabilities, offering a narrow swath with higher spatial resolution, or push-broom imaging, covering the entire globe continuously at the expense of spatial resolution. Worldwide coverage is provided by \acrshort{s5p}, \acrshort{goes}, \acrshort{gosat}, Sentinel 3, Sentinel 2 and Landsat 8/9. The not-mentioned satellites provide pointing capabilities that need specific tasking and, as a result, do not consistently monitor the exact location on Earth continuously. 

The \textit{organizations} of the second column in Table \ref{tab:all_sensors} are \acrfull{cnsa}, \acrfull{cresda}, \acrfull{cast}, \acrfull{esa}, \acrfull{noaa}, \acrfull{nasa}, \acrfull{jpl}, \acrfull{jaxa}, \acrfull{nzsa}, \acrfull{edf},  \acrfull{asi}, \acrfull{dlr}, \acrfull{usgs}. The \textit{\acrfull{eol}} column indicates the planned termination of the mission, typically representing the minimum required lifetime, which is often extended by a few years. The column \textit{\acrshort{gsd}} describes the ground dimensions of one pixel looking directly at the Earth (nadir). The \textit{Swath} describes the covered ground distance vertical to the Azimuth (Flight line). The provided \textit{revisit time} describes the time necessary to measure a specific location a second time, which correlates especially with the constellation size of the different missions and the satellites' pointing capabilities. The \textit{methane band} denotes the potential spectral ranges of sensors that could theoretically be utilized for methane monitoring, even though such utilization might not have been realized so far or is less feasible or impractical compared to other methane bands. From these spectral bands, the \textit{spectral resolution} describes the difference between the upper- and lower bound of the \acrfull{fwhm} (or best approximation) or the spectral sampling in the case of hyperspectral instruments. It is worth mentioning that some satellites like GHGSat provide multiple spectral bands in the 1650 nm area. The \textit{product levels} typically encompass the top of the atmosphere (L1), methane concentration per pixel or enhancement maps (L2), methane plume mask (L3), and emission rate/emission location (L4). The \textit{data policy} is named as governmental (gov), open access (open), commercial (com), or scientific use (sci).

While most satellites are not designed explicitly for methane measurements, multiple satellites offer Level 2 methane data, facilitating access to data for further analysis and \acrshort{ml} applications.

\scriptsize
\begin{landscape}
\begin{longtable}{lllllllllll} 
\caption{Satellite Overview sorted by GSD}
\label{tab:all_sensors}\\
\hline
\multicolumn{1}{l}{\begin{tabular}[l]{@{}l@{}} Mission\\ /Instrument\end{tabular}} & \multicolumn{1}{l}{Organization} & \multicolumn{1}{l}{\begin{tabular}[l]{@{}l@{}}lauch date\\ (EOL)\end{tabular}} & \multicolumn{1}{l}{\begin{tabular}[l]{@{}l@{}}GSD\\ (Nadir)\end{tabular}} & \multicolumn{1}{l}{\begin{tabular}[l]{@{}l@{}} Swath \\ (km) \end{tabular}} & \multicolumn{1}{l}{\begin{tabular}[l]{@{}l@{}}revisit\\ time\end{tabular}} & \begin{tabular}[l]{@{}l@{}}Methane\\ band (nm)\end{tabular} & \begin{tabular}[l]{@{}l@{}}spectral\\ resol. (nm)\end{tabular} & \begin{tabular}[l]{@{}l@{}}data\\  policy\end{tabular} & Product & Ref \\ 
\hline
\endfirsthead
\caption{Satellite Overview sorted by GSD (continued)}\\
\hline
\multicolumn{1}{l}{\begin{tabular}[l]{@{}l@{}} Mission \\ /Instrument\end{tabular}} & \multicolumn{1}{l}{Organization} & \multicolumn{1}{l}{\begin{tabular}[l]{@{}l@{}}lauch date\\ (EOL)\end{tabular}} & \multicolumn{1}{l}{\begin{tabular}[l]{@{}l@{}}GSD\\ (Nadir)\end{tabular}} & \multicolumn{1}{l}{\begin{tabular}[l]{@{}l@{}} Swath \\ (km) \end{tabular}} & \multicolumn{1}{l}{\begin{tabular}[l]{@{}l@{}}revisit\\ time\end{tabular}} & \begin{tabular}[l]{@{}l@{}}Methane\\ band (nm)\end{tabular} & \begin{tabular}[l]{@{}l@{}}spectral\\ resol. (nm)\end{tabular} & \begin{tabular}[l]{@{}l@{}}data\\  policy\end{tabular} & Product & Ref \\
\hline
\endhead
\hline

\endfoot

\endlastfoot
\multicolumn{1}{l}{\begin{tabular}[l]{@{}l@{}}Gaofen 5 \\ /\acrshort{gmi} \end{tabular}} & \multicolumn{1}{l}{\acrshort{cnsa}}  & \multicolumn{1}{l}{\begin{tabular}[l]{@{}l@{}}2021 (2028)\\ 2022 (2030)\end{tabular}}  & 10.5 km & 750 & 5 d & 1650 & 16 & gov & L1 & \multicolumn{1}{l}{\begin{tabular}[l]{@{}l@{}}  \cite{ge2022exploring}\\  \cite{ye2023improving}\end{tabular}} \\ 
\hdashline
\multicolumn{1}{l}{\begin{tabular}[l]{@{}l@{}}Huanjing 2 \\ /\acrshort{posp} \end{tabular}} & \multicolumn{1}{l}{\begin{tabular}[l]{@{}l@{}} \acrshort{cresda}, \\ CAST \end{tabular}} & 2020 (2025) & 6 km & 800 & 4 d & \multicolumn{1}{l}{\begin{tabular}[l]{@{}l@{}} 1650 \\ 2300 \end{tabular}}  & -  & gov & L1 & \multicolumn{1}{l}{\begin{tabular}[l]{@{}l@{}}  \cite{sherwin2024single}\\  \cite{dubovik2019polarimetric}\end{tabular}} \\  
\hdashline
\multicolumn{1}{l}{\begin{tabular}[l]{@{}l@{}}  \acrshort{s5p} \\ /\acrshort{tropomi} \end{tabular}} & \multicolumn{1}{l}{\acrshort{esa}} & \multicolumn{1}{l}{2017 (2027)} & \multicolumn{1}{l}{\begin{tabular}[l]{@{}l@{}} 5.5 x \\ 7 km \end{tabular}} & \multicolumn{1}{l}{2600} & daily & 2300 & 0.25 & open & L2 & \cite{fletcher2012sentinel5p} \\ 
\hdashline
\multicolumn{1}{l}{\begin{tabular}[l]{@{}l@{}} \acrshort{goes} \\ /\acrshort{abi} \end{tabular}} & \multicolumn{1}{l}{\begin{tabular}[l]{@{}l@{}}\acrshort{noaa},\\ \acrshort{nasa}\end{tabular}} & \multicolumn{1}{l}{\begin{tabular}[l]{@{}l@{}}2018 (2028)\\ 2022 (2035)\end{tabular}} & \multicolumn{1}{l}{1-2 km} & \multicolumn{1}{l}{\begin{tabular}[l]{@{}l@{}}Western \\ Hemisphere\end{tabular}} & \textless 1 d & \multicolumn{1}{l}{\begin{tabular}[l]{@{}l@{}} 1650 \\ 2300 \end{tabular}} & \multicolumn{1}{l}{\begin{tabular}[l]{@{}l@{}} 4 \\ 5 \end{tabular}} & open & L2 & \multicolumn{1}{l}{\begin{tabular}[l]{@{}l@{}} \cite{schmit2018applications} \\ \cite{watine2023geostationary} \\ \cite{schmit2017closer} \end{tabular}}\\ 
\hdashline
\multicolumn{1}{l}{\begin{tabular}[l]{@{}l@{}}\acrshort{gosat} \\ /\acrshort{tanso-fts} \end{tabular}} & \multicolumn{1}{l}{\acrshort{jaxa}} & \multicolumn{1}{l}{\begin{tabular}[l]{@{}l@{}}2009 (2023)\\ 2018 (2023)\end{tabular}} & \multicolumn{1}{l}{\begin{tabular}[l]{@{}l@{}} 1500 m \\ 920 m\end{tabular}} & \multicolumn{1}{l}{\begin{tabular}[l]{@{}l@{}} 786\\ 920\end{tabular}} & \multicolumn{1}{l}{\begin{tabular}[l]{@{}l@{}} 3 d\\ 6 d\end{tabular}} & 1650 & \multicolumn{1}{l}{0.06} & open & L2 & \cite{imasu2023greenhouse} \\ 
\hdashline
\multicolumn{1}{l}{\begin{tabular}[l]{@{}l@{}} \acrshort{noaa}-20/21 \\ \acrshort{viirs} \end{tabular}} & \multicolumn{1}{l}{\begin{tabular}[l]{@{}l@{}}\acrshort{noaa}, \\ \acrshort{nasa},\\ EUMETSAT \end{tabular}} & \multicolumn{1}{l}{\begin{tabular}[l]{@{}l@{}}2017 (2027)\\ 2022 (2030)\end{tabular}} & 750 m & 3000 & \textless 1 d & \multicolumn{1}{l}{\begin{tabular}[l]{@{}l@{}} 1650 \\ 2300 \end{tabular}} & \multicolumn{1}{l}{\begin{tabular}[l]{@{}l@{}} 60 (63) \\ 54 (48) \end{tabular}} & open & L1 & \cite{de2024daily} \\ 
\hdashline
\multicolumn{1}{l}{\begin{tabular}[l]{@{}l@{}}Sentinel 3 \\ /\acrshort{slstr} \end{tabular}} & \multicolumn{1}{l}{\acrshort{esa}} & \multicolumn{1}{l}{\begin{tabular}[l]{@{}l@{}}2016 (2026)\\ 2018 (2028)\end{tabular}} & 500 m & 1400 & 1-2 d & \multicolumn{1}{l}{\begin{tabular}[l]{@{}l@{}} 1650 \\ 2300 \end{tabular}} & \multicolumn{1}{l}{\begin{tabular}[l]{@{}l@{}} 60 \\ 50 \end{tabular}} & open & L1 & \multicolumn{1}{l}{\begin{tabular}[l]{@{}l@{}}  \cite{donlon2012global}\\  \cite{fletcher2012sentinel3}\end{tabular}} \\
\hdashline
\multicolumn{1}{l}{MethaneSAT} & \multicolumn{1}{l}{\begin{tabular}[l]{@{}l@{}} \acrshort{nzsa}, \\ \acrshort{edf} \end{tabular}} & 2024 (2028) & \multicolumn{1}{l}{\begin{tabular}[l]{@{}l@{}} 100 x \\ 400 m \end{tabular}} & 260 & 3-4 d & 1650 & 0.25 & open & \multicolumn{1}{l}{\begin{tabular}[l]{@{}l@{}}L1,\\ L2, \\ L4 \\\end{tabular}} &  \multicolumn{1}{l}{\begin{tabular}[l]{@{}l@{}}  \cite{staebell2021spectral}\\  \cite{chulakadabba2023methane}\end{tabular}} \\ 
\hdashline
\multicolumn{1}{l}{\acrshort{emit}} & \multicolumn{1}{l}{\acrshort{nasa}} & 2022 (2025) & 60 m & 80 & 36.5 d & \multicolumn{1}{l}{\begin{tabular}[l]{@{}l@{}} 1650 \\ 2300 \end{tabular}} & 7.4 & open & \multicolumn{1}{l}{\begin{tabular}[l]{@{}l@{}}L1,\\ L2 \end{tabular}} & \multicolumn{1}{l}{\begin{tabular}[l]{@{}l@{}} \cite{thompson2024orbit} \\ \cite{thorpe2023attribution} \end{tabular}} \\
\hdashline
\multicolumn{1}{l}{\begin{tabular}[l]{@{}l@{}}GHGSat(D) \\ /\acrshort{waf-p} \end{tabular}} & \multicolumn{1}{l}{\begin{tabular}[l]{@{}l@{}} GHGSat \\ Inc. \end{tabular}}  & 2016 (2024) & 50 m & 15 & 14 d & 1650 & $\sim$0.1 & com & \multicolumn{1}{l}{\begin{tabular}[l]{@{}l@{}} L2, \\ L3 \\\end{tabular}} & \multicolumn{1}{l}{\begin{tabular}[l]{@{}l@{}}  \cite{varon2019satellite} \\  \cite{jervis2021ghgsat} \end{tabular}} \\
\hdashline
\multicolumn{1}{l}{\begin{tabular}[l]{@{}l@{}} \acrshort{prisma} \\ /\acrshort{hyc} \end{tabular}} & \multicolumn{1}{l}{\acrshort{asi}} & \multicolumn{1}{l}{2019 (2024)} & 31 m & 31 & 7-29 d & \multicolumn{1}{l}{\begin{tabular}[l]{@{}l@{}}1650 \\ 2300\end{tabular}} & 6.5-11 & open & L1 & \multicolumn{1}{l}{\begin{tabular}[l]{@{}l@{}}  \cite{cogliati2021prisma}\\  \cite{guanter2021mapping}\end{tabular}} \\ 
\hdashline
\multicolumn{1}{l}{\begin{tabular}[l]{@{}l@{}}GHGSat(CX) \\ /WAF-P \end{tabular}} & \multicolumn{1}{l}{\begin{tabular}[l]{@{}l@{}} GHGSat \\ Inc. \end{tabular}} & \multicolumn{1}{l}{\begin{tabular}[l]{@{}l@{}}2020 (2024)\\ multiple (2027)\end{tabular}}  & 30 m & 15 & 1-2 d & 1650 & $\sim$0.1 & com & \multicolumn{1}{l}{\begin{tabular}[l]{@{}l@{}} L2, \\ L3 \\\end{tabular}} & \multicolumn{1}{l}{\begin{tabular}[l]{@{}l@{}}  \cite{varon2019satellite}\\  \cite{jervis2021ghgsat} \end{tabular}} \\
\hdashline
\multicolumn{1}{l}{\begin{tabular}[l]{@{}l@{}}Gaofen 5 \\ /\acrshort{ahsi} \end{tabular}} & \multicolumn{1}{l}{\acrshort{cnsa}} & \multicolumn{1}{l}{\begin{tabular}[l]{@{}l@{}}2021 (2028)\\ 2022 (2030)\end{tabular}} & 30 m & 60 & 5 d & \multicolumn{1}{l}{\begin{tabular}[l]{@{}l@{}}1650 \\ 2300\end{tabular}} & 10 & gov & L1 & \cite{liu2019advanced} \\
\hdashline
\multicolumn{1}{l}{\begin{tabular}[l]{@{}l@{}}Ziyuan 1-02D/E \\ /\acrshort{ahsi} \end{tabular}} & \multicolumn{1}{l}{\acrshort{cresda}} & \multicolumn{1}{l}{\begin{tabular}[l]{@{}l@{}}2021 (-) \\ 2019 (-) \end{tabular}} & 30 m & 60 & 1-3 d & \multicolumn{1}{l}{\begin{tabular}[l]{@{}l@{}}1650 \\ 2300\end{tabular}} & 20 & gov & L1 & \multicolumn{1}{l}{\begin{tabular}[l]{@{}l@{}}  \cite{song2022quantifying}\\  \cite{sherwin2024single} \\ \cite{yang2022spectral}\end{tabular}} \\ 
\hdashline
\multicolumn{1}{l}{\begin{tabular}[l]{@{}l@{}}\acrshort{enmap} \\ /\acrshort{hsi} \end{tabular}} & \multicolumn{1}{l}{\acrshort{dlr}} & 2022 (2026) & 30 m & 30 & 4 - 27 d & \multicolumn{1}{l}{\begin{tabular}[l]{@{}l@{}}1650 \\ 2300\end{tabular}} & 10 & Sci & L1 & \multicolumn{1}{l}{\begin{tabular}[l]{@{}l@{}}  \cite{chabrillat2022enmap}\\  \cite{guanter2015enmap}\end{tabular}} \\
\hdashline
\multicolumn{1}{l}{\begin{tabular}[l]{@{}l@{}}Landsat 8/9 \\ /\acrshort{oli} \end{tabular}} & \multicolumn{1}{l}{\begin{tabular}[l]{@{}l@{}} \acrshort{nasa}, \\ \acrshort{usgs} \end{tabular}} & \multicolumn{1}{l}{\begin{tabular}[l]{@{}l@{}}2013 (2028) \\ 2021 (2031)\end{tabular}} & 30 m & 185 & $\sim$8 d & \multicolumn{1}{l}{\begin{tabular}[l]{@{}l@{}}1650 \\ 2300\end{tabular}} & \multicolumn{1}{l}{\begin{tabular}[l]{@{}l@{}}85 \\ 185\end{tabular}} & open & L1 & \multicolumn{1}{l}{\begin{tabular}[l]{@{}l@{}}  \cite{li2017global}\\  \cite{Ihlen2019Landsat8HB} \\ \cite{Sayler2022Landsat9HB}\end{tabular}} \\ 
\hdashline
\multicolumn{1}{l}{\begin{tabular}[l]{@{}l@{}}Sentinel 2 \\ /\acrshort{msi} \end{tabular}} & \multicolumn{1}{l}{\acrshort{esa}} & \multicolumn{1}{l}{\begin{tabular}[l]{@{}l@{}}2015 (2025) \\ 2017 (2027)\end{tabular}} & 20 m & 290 & \textless5 d & \multicolumn{1}{l}{\begin{tabular}[l]{@{}l@{}}1650 \\ 2300\end{tabular}} & \multicolumn{1}{l}{\begin{tabular}[l]{@{}l@{}} 90 \\ 180 \end{tabular}} & open & L1 & \multicolumn{1}{l}{\begin{tabular}[l]{@{}l@{}}  \cite{gorrono2023understanding}\\ \cite{fletcher2012sentinel2} \end{tabular}} \\
\hdashline
\multicolumn{1}{l}{\begin{tabular}[l]{@{}l@{}}\acrshort{geisatp} \\ /iSIM-90 \end{tabular}} & SATLANTIS & 2023 (2027) & 13 m & 16.5 & - & 1650 & - & com & - & \multicolumn{1}{l}{\begin{tabular}[l]{@{}l@{}}  \cite{ubierna2022gei}\\ \cite{eoportal_geisat_24}\end{tabular}} \\ 
\hdashline
\multicolumn{1}{l}{\begin{tabular}[l]{@{}l@{}}Aurora/\acrshort{ghost} \\ /\acrshort{hi} \end{tabular}} & \multicolumn{1}{l}{\begin{tabular}[l]{@{}l@{}} Orbital \\ Sidekick Inc. \end{tabular}} & \multicolumn{1}{l}{\begin{tabular}[l]{@{}l@{}} 2021 (2026) \\ 2023 (2029) \end{tabular}} & 8 m & - & \textless1 d & \multicolumn{1}{l}{\begin{tabular}[l]{@{}l@{}}1650 \\ 2300\end{tabular}} & $\sim$ 4.1 & com & - & \multicolumn{1}{l}{\begin{tabular}[l]{@{}l@{}}  \cite{WMO_GHO_24}\\ \cite{CEOS_GHO_24}\end{tabular}} \\
\hdashline
\multicolumn{1}{l}{WorldView-3} & \multicolumn{1}{l}{\begin{tabular}[l]{@{}l@{}} Maxar \\ Technologies \\ Inc. \end{tabular}}  & 2014 (2024) & 3.7 m & 13.1 & $\sim$1 d & \multicolumn{1}{l}{\begin{tabular}[l]{@{}l@{}}1650 \\ 2300\end{tabular}} & \multicolumn{1}{l}{\begin{tabular}[l]{@{}l@{}}40 \\ 50, 70\end{tabular}} & com & L1 & \cite{sanchez2022mapping}  \\
\hdashline
\multicolumn{1}{l}{\begin{tabular}[l]{@{}l@{}}\acrshort{avirisng}\\ /(Air)\end{tabular}} & \multicolumn{1}{l}{\begin{tabular}[l]{@{}l@{}}\acrshort{nasa},\\ \acrshort{jpl}\end{tabular}} & 2014 & $\sim$3 m & $\sim$1.85 & airborne & \multicolumn{1}{l}{\begin{tabular}[l]{@{}l@{}}1650 \\ 2300\end{tabular}} & 5 & open & L1 & \multicolumn{1}{l}{\begin{tabular}[l]{@{}l@{}}\cite{cusworth2019potential} \\ \cite{thorpe2016mapping}\\ \cite{thorpe2014retrieval} \end{tabular}} \\
\hdashline
\multicolumn{1}{l}{\begin{tabular}[l]{@{}l@{}}\acrshort{aviris}-3\\ /(Air)\end{tabular}} & \multicolumn{1}{l}{\begin{tabular}[l]{@{}l@{}}\acrshort{nasa},\\ \acrshort{jpl}\end{tabular}} & 2023 & $\sim$2 m & $\sim$2.1 & airborne & \multicolumn{1}{l}{\begin{tabular}[l]{@{}l@{}}1650 \\ 2300\end{tabular}} & 7.5 & open & L1 & \cite{green2022airborne} \\
\hline
\end{longtable}
\end{landscape}
\normalsize

Some satellites mentioned in Table \ref{tab:all_sensors} provide limitations or changes that can not be assessed from the Table. 
 \acrshort{s5p}'s resolution changed from 7.5 km x 7.5 km to 5.5 km x 7.5 km on 06. August 2019. 
The specifications of the satellites \acrshort{noaa}-20 and \acrshort{noaa}-21 differ slightly, so the spectral resolution in Table \ref{tab:all_sensors} shows the resolution of \acrshort{noaa}-21 in parentheses. 
The \acrshort{emit} instrument is mounted on the \acrshort{iss}, limiting the mapping area from 51.6° to -51.6° latitude and providing a long revisit time of 36.5 days for some points.
MethaneSAT has just been launched and is still in the commissioning phase at the time of writing. This is why the processing levels are estimated, and the operational instrument performance has not been proven yet. 
Landsat 8 and Landsat 9 are different satellites, but the \acrfull{oli} instruments provide similar optics, which is why the data is similar, and algorithms can be applied for both. 
The Satellites Gaofen 5, Huanjing 2, and Ziyuan 1 are usually just for the organizations' use. However, the satellites are included in this list due to a more open data policy for the work of \cite{sherwin2024single}. 
The GHGSat-CX constellation consists of the satellites GHGSat-C1 to GHGSat-C10 for methane monitoring, providing a high revisit time through the number of satellites and pointing capabilities. 
Due to limited publicly available information, some information is missing for the commercial missions \acrfull{geisatp} and Aurora / \acrfull{ghost}. 
The mission \acrshort{goes} refers to \acrshort{goes}-16 (East) and \acrshort{goes}-18 (West), with \acrshort{goes}-17 as an on-orbit standby satellite, which are the only geostationary satellites for methane monitoring so far. These satellites provide coverage from 150° to 0° longitude.
Worldview-3 provides multiple bands sensitive to methane, two of which provide different spectral resolutions in the 2300 nm band; one band provides 50 nm spectral resolution and another one 70 nm. 
Both \acrshort{aviris} sensors provide a \acrshort{gsd} of 0.3 to 20 m depending on the flight height, with a sampling of 0.6 milliradian for \acrshort{aviris}-3 and 1 milliradian for \acrshort{avirisng}. The Swath Angle of \acrshort{aviris}-3 and \acrshort{avirisng} is 40° and 34°, respectively \citep{green2022airborne}. With the typical flight height of 3 to 4 km from more extensive campaigns \citep{frankenberg2016airborne, duren2019california}, we assumed a 3300 m flight height resulting in a \acrshort{gsd} of about 3.3 m and Swath of 1.8 km for \acrshort{avirisng} and 2 m \acrshort{gsd} and 2.1 km swath for \acrshort{aviris}-3.

As for the products provided by these systems, the main focus is not on targeting methane observations due to the vast availability of Level 1 data. Only  \acrshort{s5p}, \acrshort{goes}, and \acrshort{gosat}, as well as the Methane-focused missions of MethaneSAT and GHGSat, provide concentration maps that can be used for methane observations. MethaneSAT and GHGSat even provide higher-level methane products using some approaches mentioned hereafter. 
Even though most satellites provide a methane band of 1650 nm, the predominantly used methane band is 2300 nm as it yields improved results, explained in Chapter \ref{chap:Physical-based method_general}. 
Commercial systems that need to be purchased generally provide a higher revisit time and spatial resolutions due to pointing capabilities than publicly available systems. This reflects the detection limits as described in Chapter \ref{chap:detection_thresholds}. 
The downside of free available systems is either the revisit time (through fewer satellites), \acrshort{gsd}, or spectral resolution.
GHGSat is the only system that provides a very small bandwidth for the methane-sensitive bands at 1650 nm, with about 0.1 nm providing excellent spatial resolution.

\subsection{Future Satellites for methane detection}
\label{chap:future_systems}

The increasing attention in recent years for methane monitoring has resulted in multiple satellite missions being able to fulfill this task. In Table \ref{tab:future_ch4_satellites}, we provide an overview of the evolving landscape of satellite-based methane monitoring, focusing on the latest developments in satellite technology and upcoming missions planned for launch. Here, we concentrate on satellite missions, which provide basic information about the instruments used and their performance. In contrast to the mentioned future methane missions in \citep{jacob2022quantifying}, the Geocarb mission has been excluded due to cancellation announced by \acrshort{nasa} in 2022 \citep{cancel_geocarb_22}. Similarly, the Microcarb mission strongly focuses on \acrshort{co2} but removed the methane bands during development \citep{eoportal_microcarb_24}. The systems mentioned in Table \ref{tab:future_ch4_satellites} and the existing systems all rely on the solar backscatter of the earth to measure methane. However, the \acrfull{merlin} instrument is an exception as it emits its own signal and analyses the reflection of the signal to measure methane \citep{ehret2017merlin}. 
Table \ref{tab:future_ch4_satellites} describes the \textit{launch date} as a planned launch or launch dates of multiple satellites for each mission, the \textit{methane band} is the band which will be primarily used for methane detection, \textit{spect resol} is the spectral resolution defined as the \acrshort{fwhm} or provided spectral resolution. The columns \textit{revisit time} and \textit{\acrshort{gsd}} align with the existing sensors in Table \ref{tab:all_sensors}. Particular information that is not publicly available has been marked as ``\textbf{-}".

\begin{table}[h]
    \caption{Future Satellites for methane monitoring}    
    \centering
    \resizebox{\textwidth}{!}{
    \begin{tabular}{llllllll}
            \hline
            Mission & \multicolumn{1}{l}{\begin{tabular}[l]{@{}l@{}} Launch \\ date \end{tabular}} & \multicolumn{1}{l}{\begin{tabular}[l]{@{}l@{}} Methane \\ band (nm) \end{tabular}} & \multicolumn{1}{l}{\begin{tabular}[l]{@{}l@{}} spect resol \\ in (nm) \end{tabular}}  & \multicolumn{1}{l}{\begin{tabular}[l]{@{}l@{}} revisit \\ time (d) \end{tabular}}  & GSD  & Ref  \\ 
            \hline
            Carbon Mapper & 2024 & 2300 & 6 & 1-7 & \multicolumn{1}{l}{\begin{tabular}[l]{@{}l@{}} 30x30m$^2$ \\ 30x60m$^2$ \end{tabular}} & \cite{duren2021carbon} \\ 
            \acrshort{gosatgw} & 2024 & 1650 & 0.2 & 3 & \multicolumn{1}{l}{\begin{tabular}[l]{@{}l@{}} 10km (wide) \\ 1-3km (focus) \end{tabular}} &\cite{GOSATGW_website24} \\ 
            BrightSkies & 2024 & 1650 & \textless 0.5 & 4 & 100m &  \cite{CEOS_BrightSkies_24} \\ 
            \acrshort{tango} & 2026 & 1650 & 0.45 & - & 300m &  \cite{brenny2023development} \\ 
            TanSat-2 & 2025 & 2300 & 0.11 & 2-5 & 2x2km  & \cite{CEOS_Tansat2_24} \\ 
            GeiSAT (iSIM-170) & 2025 & 2300 & - & - & 9m & \cite{ubierna2022gei} \\ 
            \acrshort{co2m} & 2025/26/27 & 1650 & 0.3 & 5 & 2x2km$^2$  & \cite{sierk2021copernicus} \\ 
            Sentinel 5 & 2025/31/38 & both & 0.25 & daily & 7.5x7.5km$^2$ & \cite{irizar2019sentinel} \\ 
            \acrshort{merlin} & 2027 & 1650 & 3x10$^{-4}$ & 28 & up to 150m$^2$ &\cite{ehret2017merlin} \\
            \acrshort{chime} & 2028/30 & both & \textless 10 & 10-12.5 & 20-30m & \cite{candiani2022evaluation} \\ 
            \hline
            \end{tabular}
    }
    \label{tab:future_ch4_satellites}
\end{table}

The GeiSAT instrument, iSIM-170, was already demonstrated in 2020 during a test mounted on the \acrshort{iss}. \citep{ubierna2022gei}. The private company GHGSat has planned to launch four more GHGSat-CX satellites, which are not listed here, as they will carry the same instrument as their predecessors. 
For the same reason, the planned \acrshort{goes}-19 satellite is not listed here. It carries the same \acrfull{abi} as the existing \acrshort{goes} satellites. 
Carbon mapper and \acrfull{gosatgw} provide two capturing modes, one with a focus and one wider for different use cases, which results in the two different \acrshort{gsd}s.
The future missions mentioned here show different satellites for large spatial areas and point emissions, building a new and improved baseline for diverse methane detection tasks.

The future missions tend to focus on higher spectral resolution instead of spatial resolution due to the higher spectral resolution compared to existing systems. This may limit the capabilities to detect small plumes but should increase the concentration estimation and potentially the emission rate estimation in the end.

\subsection{Comparison of existing sensors for methane detection}
\label{chap:detection_thresholds}

The sensors discussed in the preceding parts of Chapters \ref{chap:satellite_systems} offer diverse functionalities, applications, and capabilities. The primary focus of this section revolves around the detection thresholds of these systems, representing the minimum plume size that can still be detected and quantified in a good-case scenario. This parameter is often called the \textit{minimum point source emission rate} in the literature.

Aligned with a previous study of \cite{jacob2022quantifying}, we have adopted certain detection thresholds as well as refined thresholds based on recent research and included systems not mentioned previously. Due to incomplete information regarding the detection thresholds of specific systems, not all future systems could be integrated into the comparative analysis presented in Table \ref{tab:sensors_thresholds}.

Satellites such as  \acrshort{s5p} or \acrshort{gosat} are predominantly employed for monitoring country or worldwide emissions at a large scale, benefiting from their high spatial coverage, which results in a high detection threshold. Additionally, systems like \acrshort{gosat} are utilized to establish monthly or yearly methane emission budgets for large areas. This differs from their application in estimating emission rates of individual plumes. 
The detection threshold of \acrshort{gosat} for single-source emission has not been investigated intensively by the scientific community. Therefore, the threshold has been calculated through the theoretical analysis of \acrshort{gosat} and  \acrshort{s5p} of 7.1 and 4.2 tons per hour \citep{jacob2016satellite}. Through this relation, we assumed a higher detection threshold for \acrshort{gosat} as for  \acrshort{s5p}, which was investigated by \cite{lauvaux2022global}.

Sentinel 2 and Landsat 8/9's detection thresholds have been adopted to new research by \cite{gorrono2023understanding} stating a more accurate detection threshold of 1000-5000 kg/h.

Notably, most of these systems were not primarily designed for methane monitoring or point source emissions but provided satisfactory results. Only GHGSat and methaneSAT are precisely engineered with a focus on methane measurement.

Due to the limited access to the Chinese governmental satellite missions Ziyuan 1 and Gaofen 5, public analysis of methane detection thresholds could not be found during writing. Usually, these satellites are only available to the Chinese government. However, in recent work by \cite{sherwin2024single}, three research groups received data on these satellites for comparison. The authors found that both satellites could detect plumes in actual observations with an emission rate of about 1000 kg/h, mentioning that the satellites could detect emissions lower under favorable conditions.

Generally, a lower detection threshold is observed for satellites with lower \acrshort{gsd}. Hyperspectral satellites offer superior spectral properties and narrower bands, which proves advantageous in detecting smaller plumes. Dedicated methane detection systems like GHGSat feature very narrow bands in the spectral absorption range of methane (1650 nm), resulting in the lowest detection thresholds achieved by satellites thus far.

\begin{table}[h]
    \caption{Methane emission rate thresholds for detection of researched systems}    
    \centering
    \resizebox{\textwidth}{!}{
    \begin{tabular}{lllll}
            \hline
            Mission & \multicolumn{1}{l}{\begin{tabular}[l]{@{}l@{}} Detection \\ threshold (kg/h) \end{tabular}} & \multicolumn{1}{l}{\begin{tabular}[l]{@{}l@{}} GSD \\ (nadir) \end{tabular}} & type & Ref \\   
            \hline
            \acrshort{gosat} & $>$25000 & 1500 m & multi & \cite{jacob2016satellite} \\ 
             \acrshort{s5p} & 25000 & 5.5x7.5 km & multi & \cite{lauvaux2022global} \\
            Sentinel 3 & 8000 - 20000 & 500 m & multi & \cite{pandey2023daily} \\
            \acrshort{noaa}-20/21 & 8000 - 20000 & 750 m & multi & \cite{de2024daily} \\
            Sentinel 2 & 1000 - 5000 & 20 m & multi & \multicolumn{1}{l}{\begin{tabular}[l]{@{}l@{}} \cite{gorrono2023understanding}, \\ \cite{ehret2022global} \end{tabular}} \\ 
            Landsat 8/9 & 1000 - 5000 & 30 m & multi & \multicolumn{1}{l}{\begin{tabular}[l]{@{}l@{}} \cite{gorrono2023understanding}, \\ \cite{ehret2022global} \end{tabular}} \\ 
            GHGSat-D & 1000 - 3000 & 50 m & multi & \cite{jervis2021ghgsat} \\
            \acrshort{prisma} & 500 - 2000 & 30 m & hyp & \cite{guanter2021mapping} \\
            Ziyuan 1 & \textless 1000 & 30 m & hyp & \cite{sherwin2024single} \\
            Gaofen 5 & \textless 1000 & 30 m & hyp & \cite{sherwin2024single} \\
            \acrshort{emit} & 200 - 300 & 60 m & hyp & \cite{thorpe2023attribution} \\
            \acrshort{enmap} & 100 - 500 & 30 m & hyp & \cite{cusworth2019potential} \\
            GHGSat-CX & 100 - 200 & 30 m & multi & \cite{gauthier2021importance} \\
            \acrshort{geisatp} (iSIM-90) & $\sim$150 & 13 m & multi & \cite{eoportal_geisat_24} \\ 
            WorldView 3 & \textless 100 & 3.7 m & multi & \cite{sanchez2022mapping} \\
            \multicolumn{1}{l}{\begin{tabular}[l]{@{}l@{}} \acrshort{avirisng} \\ (airborne) \end{tabular}}  & 2 - 10 & $\sim$3 m & hyp & \multicolumn{1}{l}{\begin{tabular}[l]{@{}l@{}} \cite{duren2019california}, \\ \cite{thorpe2020methane} \end{tabular}} \\
            \multicolumn{1}{l}{\begin{tabular}[l]{@{}l@{}} \acrshort{aviris}-3 \\ (airborne) \end{tabular}} & \textless 10 & $\sim$2 m & hyp & \cite{coleman2024quantification} \\
            \hline
            \acrshort{tango} & 10000 & 100 m & multi & \cite{brenny2023development} \\
            MethaneSAT & 500 & 100x400 m & multi & \cite{jacob2022quantifying} \\
            BrightSkies & 100 & 100 m & multi & \cite{CEOS_BrightSkies_24} \\ 
            Carbon Mapper & 50 - 200 & 30 m & multi & \cite{duren2021carbon} \\
            GeiSat (iSIM-170) & $\sim$50 & 9 m & multi & \cite{eoportal_geisat_24} \\
            \hline
            \end{tabular}
    } 
    \label{tab:sensors_thresholds} 
\end{table}

While the airborne sensor \acrshort{aviris}-3 is assumed to provide a better detection threshold to \acrshort{avirisng}, it is important to note that no work has yet analyzed the actual detection limit of this sensor. However, ongoing research on the detection capabilities of \acrshort{aviris}-3, led by RW Coleman and other researchers \citep{coleman2024quantification}, holds the potential for future discoveries in methane detection technology. 

The future of satellite systems (i.e. the last 5 missions in Table \ref{tab:sensors_thresholds}) holds great promise. These advancements are about achieving even lower detection thresholds and signify expanding our capabilities to detect increasingly smaller methane plumes. This potential for future advancements should inspire optimism and excitement in environmental science and satellite technology.

\section{Traditional approaches for optical satellite-based methane plume detection}
\label{chap:traditional_approaches}

Methane can be detected using three primary types of platforms: ground-based, airborne, and satellite-based systems. Among them, satellite-based detection can provide stable and continuous global monitoring of methane emissions \citep{jacob2016satellite, jacob2022quantifying}. Methane's selective radiation absorption at approximately 1650 nm and 2300 nm results in distinctive spectral absorption features. Consequently, the retrieval of methane concentrations from satellite observations typically relies on spectrally-resolved measurements of solar radiation reflected by the Earth's surface within the \acrshort{swir} region of the spectrum, spanning approximately 1600 to 2500 nm \citep{worden2015quantifying, guanter2021mapping}. The previously mentioned satellite systems have been used for methane observations, with multiple more to be launched. This Chapter describes the approaches for detecting and quantifying methane from satellite observations. 

\subsection{Methane retrieval}
The Methane retrieval process aims to determine the enhancement of methane column concentration per pixel with respect to the background (\(\Delta\)XCH\textsubscript{4}), mainly divided into physically-based and statistical methods. Physical-based methodologies explicitly model radiative transfer interactions among the surface, atmosphere, and instrument. Conversely, statistical methods are utilized to extract relevant information from the image. 

\subsubsection{Physical-based method}
\label{chap:Physical-based method_general} 
Estimating methane concentration enhancement typically involves fitting high-spectral-resolution observations in the \acrshort{swir} spectral region to a modeled radiance spectrum during a single satellite overpass \citep{thorpe2014retrieval,jacob2016satellite}. 
Physical-based methane concentration enhancement map retrieval involves two primary methods: the \textit{full-physics} retrieval and the \textit{\acrshort{co2} proxy retrieval}. 

The \textit{full-physics} approach utilizes a radiative transfer model to invert spectra, aiming to simultaneously solve for vertical profiles of methane concentration, aerosol extinction, and surface reflectivity. For this approach the percentage of good estimates of methane concentration per pixel (success rate) using TROPOMI data in the 2300 nm band is only 3\% over land \citep{jacob2022quantifying, lorente2021methane}. Besides, this method faces challenges in capturing vertical gradients and is heavily influenced by atmospheric conditions and surface heterogeneity \citep{jacob2022quantifying}.  
Conversely, the \textit{\acrshort{co2} proxy retrieval} leverages the adjacent \acrshort{co2} absorption band at 1610 nm to estimate methane and \acrshort{co2} concentrations simultaneously \citep{frankenberg2005iterative}, achieving similar precision and accuracy \citep{buchwitz2015greenhouse}. Despite being faster and less affected by surface and aerosol biases, the method using the \acrshort{gosat} instrument at 1650 nm still encounters errors in regions with unresolved \acrshort{co2} variability and from sources co-emitting methane and \acrshort{co2} like flares, achieving a 24\% success rate over land applied on GOSAT data, primarily limited by cloud cover \citep{parker2020decade}. 
Research has highlighted that higher spectral resolution enables more precise measurements, as demonstrated by \cite{cusworth2019potential} and \cite{jongaramrungruang2021remote}. Additionally, the positioning of spectral bands relative to methane absorption lines significantly influences precision and accuracy, as quantified by \cite{jacob2022quantifying}. 
Hyperspectral data allows for joint optimization of methane, other trace gases, and surface albedo from a single observation \citep{jacob2016satellite,cusworth2019potential,irakulis2021satellite,guanter2021mapping,roger2024high}, multispectral instruments with broadband channels, despite their lower spectral resolution (typically around 100 nm), can also effectively retrieve methane column concentration enhancements. 
This is done using two spectral measurements: one that includes methane emissions and one that does not. Dataset recorded by Sentinel 2 has been proved to successfully estimate methane column enhancements in a plume relative to the background by inferring surface reflectivity from adjacent bands around 1600 and 2300 nm, or from observations of the same scene when the plume is absent  \citep{varon2021high,sanchez2022mapping,ehret2022global,gorrono2023understanding}. The main principles could be concluded as below \citep{varon2021high}: 
\begin{itemize}
    \item The \acrfull{sbmp}  retrieval method\\
    The \acrshort{sbmp} retrieval method compares TOA reflectances from the spectral band (located around 2300 nm) over a methane source to those measured without emissions. It derives methane concentration enhancements from the fractional change in reflectance, adjusted by a scaling factor, and uses a Gauss–Newton method for retrieval \citep{varon2021high}. While conceptually simple, \acrshort{sbmp} requires multiple satellite passes and may struggle with identifying plume-free passes for persistent methane sources. It is also sensitive to non-uniform changes in surface albedo over time. However, background water vapor and \acrshort{co2} variations have minimal effect on the retrieval process \citep{duren2019california}.
    
    \item The \acrfull{mbsp}  retrieval method\\
    The \acrshort{mbsp} retrieval method estimates methane enhancements by comparing reflectances from two adjacent spectral bands (ideally located at 1650 and 2300 nm) during a single satellite pass. It utilizes least-squares fitting to determine the fractional change in reflectance and employs a fractional absorption model considering methane sensitivity. \acrshort{mbsp} offers the advantage of requiring only one satellite pass for methane concentration retrieval but relies on signals from spectral bands with central wavelengths separated by 600 nm, is too wide a gap and could lead to errors due to varying aerosol reflectance properties in the two bands. 
    \item The \acrfull{mbmp}  retrieval method\\    
    The third retrieval method integrates aspects of the first two, determining methane column enhancements by comparing \acrshort{mbsp} retrievals from distinct satellite passes. Here, systematic errors in the \acrshort{mbsp} retrieval (\acrshort{mbsp}) caused by wavelength separation between bands are rectified by subtracting another \acrshort{mbsp} retrieval (\acrshort{mbsp}') conducted during a satellite pass devoid of methane plume occurrences. This correction aims to eliminate artifacts inherent in the retrieval field, preserving only genuine methane enhancements if systematic errors in the \acrshort{mbsp} retrievals remain consistent across both passes. 
\end{itemize}
The \acrshort{mbmp} retrieval method typically achieves superior precision and lower plume detection limits across various scene types. However, in situations with variable surface conditions, where defining a reference is challenging, the single-pass \acrshort{mbsp} method shows slightly better performance \citep{varon2021high}.

\subsubsection{Statistical methods} 
Statistical methods offer an alternative approach to methane retrieval by utilizing statistical techniques to constrain the retrieval process with information extracted directly from the image. They comprise methods such as matched-filter and singular vector decomposition concepts and have shown success with imaging spectroscopy data \citep{thorpe2014retrieval, thompson2016space, foote2020fast, guanter2021mapping, roger2024high} and multispectral satellite images \citep{wang2024exploiting}. One of the primary advantages of statistical retrievals is their ability to implicitly account for potential radiometric and spectral errors commonly found in satellite imaging spectroscopy data, such as vertical striping from detector non-uniformity \citep{guanter2021mapping}. 
In contrast to the physical based model, the statistical models provide (\(\Delta\)XCH\textsubscript{4}) estimates directly without requiring additional background considerations, unlike physically based methods.
Additionally, statistical retrievals show substantially superior computational efficiency compared to physically based methods, making them an efficient solution for methane retrieval. 

Matched filters are the most used method and have proven to be effective for retrieving methane enhancements from point sources. This technique entails aligning the observed spectrum with a background spectrum that has been convolved with a target methane absorption spectrum, specifically targeting the 2300 nm absorption band. Researchers have extensively applied matched filter methods to identify methane emissions from various sources, including industrial facilities and natural gas infrastructure, utilizing \acrshort{aviris} data from studies conducted by \cite{frankenberg2016airborne}, \cite{duren2019california}, and \cite{cusworth2021intermittency}. Furthermore, matched-filter methods have been successfully adapted for satellite retrieval of point sources, as shown in studies conducted by \cite{thompson2016space}, \cite{guanter2021mapping}, and \cite{irakulis2021satellite}. A vital advantage of these methods is their ability to directly retrieve methane enhancement above the background, offering a quicker alternative to \textit{full-physics} retrieval methods. This efficiency makes matched-filter methods particularly suitable for plume imaging applications, where capturing methane enhancement above the local background is of primary interest. 

This technique models the background radiance using a multivariate Gaussian distribution characterized by its mean value (\(\mu\)) and covariance matrix (\(\Sigma\)). Deviations from the modeled background radiance indicate \acrshort{ch4} concentration enhancements. 
\begin{equation}
 x = \mu + \Delta X_{\text{CH}_4} \cdot t 
\end{equation}
The formula relates the at-sensor radiance spectrum (\(x\)) to the methane concentration enhancement. The target signature \( t \) is obtained by multiplying the mean background radiance (\(\mu\)) with the unit methane absorption spectrum (\(k\)). The \(k\) spectrum is computed using the \acrfull{modtran} radiative transfer model, which simulates methane transmittance spectra for different mixing ratios. Calculating \(k\) requires setting \acrshort{modtran} to spectral transmittance mode and applying atmospheric profiles from the U.S. Standard Atmosphere. The fitting process correlates methane enhancements with transmittance changes, and derivatives quantify the absorption corresponding to a unit methane concentration.  
The calculation integrates over an 8 km vertical column for satellite observations \citep{thompson2016space}. This method is often employed in the 2100–2450 nm spectral range to detect methane. Despite higher radiance levels typically observed in the 1650 nm absorption window, its weaker absorption and fewer spectral bands covering this range lead to noisier methane retrievals \citep{roger2024high}. Although, matched-filters can be used to calculate methane concentration for each pixel within a higher spatial resolution compared to area mappers, on water, low albedo surfaces, and other challenging conditions it still tend to have lower success rate.

\subsection{Emission rate estimation}
\label{chap:review_emission_rate_traditional}

Methane concentrations retrieved via remote sensing are normally described using the column-average dry molar mixing ratio, symbolized as \(X\) [ppb]. It characterizes the methane plume by the enhancement \(\Delta X\), which is calculated as the deviation of \(X\) from the local background concentration \(X_b\). To correlate plume observations with the emission source rate \(Q\) [kg h\(^{-1}\)], the column mass enhancement \(\Delta \Omega\) [kg m\(^{-2}\)] is a more practical measure. The column mass enhancement \(\Delta \Omega\) is related to the enhancement \(\Delta X\) through the formula:

\begin{equation}
    \Delta \Omega = \frac{M_{\text{CH}_4}}{M_a} \Omega_a \Delta X 
\end{equation}

In this equation, \(M_{\text{CH}_4}\) and \(M_a\) represent the molar masses of methane and dry air, respectively, in units of kg mol\(^{-1}\). The variable \(\Omega_a\) signifies the column amount of dry air, measured in units of kg m\(^{-2}\). 

Quantifying point source emissions from satellite observations of instantaneous methane plumes involves a unique inversion challenge. The main objective is to determine the emission rate from a single snapshot of the plume. While the plume's morphology is influenced by turbulent diffusion and mean wind, the observation typically captures the total methane column, thus minimizing errors related to vertical boundary layer mixing. However, the absence of precise wind speed information poses a significant challenge, as variations in wind speed directly impact concentration ratios in the plume, propagating errors in wind speed estimation to the inferred point source rate. Various methods have been developed to address this challenge \citep{jacob2022quantifying}, including the Gaussian plume model, mass balance method, Gauss theorem method, cross-sectional flux (CSF) method, and \acrshort{ime} method. The \textit{CSF} and \textit{\acrshort{ime}} methods are most frequently used to derive point source rates from satellite observations, consistently producing reliable results \citep{krings2011mamap,krings2013quantification,frankenberg2016airborne,varon2018quantifying}. 

The \textit{CSF} method evaluates emission source rates by calculating the flux in the cross-section of the plume orthogonal to the plume axis, initially applied in aircraft \textit{in situ} observations \citep{varon2018quantifying} and adapted for remote sensing datasets \citep{krings2011mamap, tratt2014airborne}. While remote sensing offers comprehensive vertical coverage compared to \textit{in situ} methods, it lacks detailed wind characterization, relying on an average vertical wind speed \( U_{eff} \) parameterized from the 10-meter wind speed \citep{varon2018quantifying}  or interpolated vertical profiles \citep{krings2011mamap}. The emission rate Q is given by: 

\begin{equation}
    \mathrm{Q} = U_{eff} \int_{-\infty}^{+\infty}\Delta \Omega(x, y) \, dy 
\end{equation}

The \textit{\acrshort{ime}} method connects the source rate with the total mass of the plume detected downwind, providing a quantitative approach to assess emissions. To calculate the \acrshort{ime} of an observed column plume, which consists of \( N \) pixels, the sum of the product of the column amount for each pixel and its respective area $A_j$ is computed:

\begin{equation}
    \mathrm{IME} = \sum_{j=1}^{N}\Delta \Omega_j A_j 
\end{equation}

\cite{frankenberg2016airborne} established an empirical linear relationship between \acrshort{ime} and the source rate Q for methane plumes detected in airborne data, utilizing independent estimates from the cross-sectional flux method to define this relationship \citep{frankenberg2016airborne}. More fundamentally, the connection between \acrshort{ime} and Q is determined by the residence time \( \tau \) of methane in the detectable plume. The residence time \( \tau \) can be dimensionally represented using an effective wind speed \( U_{eff} \) and a characteristic plume size \( L \):

\begin{equation}
    \mathrm{Q} = \frac{1}{\tau} \mathrm{IME} = \frac{U_{eff}}{L} \mathrm{IME}
\end{equation}

Under idealized conditions, \( U_{eff} \) and \( L \) would directly correspond to the wind speed and plume length, assuming uniform transport to a terminal distance. However, in reality, plume dissipation is driven by turbulent diffusion in all directions. Consequently, \( U_{eff} \) and \( L \) must be treated as operational parameters that are linked to observational data on wind speed and the spatial extent of the plume. The derivation from synthetic plumes indicates that the detectable plume size \( L \) is influenced by both Q and \( U_{eff} \), introducing non-linearity into the equation.

While the CSF method is more physically based and allows for error reduction through cross-sectional analysis, the \acrshort{ime} method parameterizes total mass enhancement in the plume relative to wind speed. Both methods require accurate wind speed estimates, typically obtained from meteorological databases or measurements at the point source location, which can dominate the error budget and limit precision to around 30\% \citep{varon2018quantifying}. 

Another important factor in calculating emission rates is that both methods require the identification and masking (i.e., segmentation) of the plume in the enhancement map. Plume detection and masking in satellite imagery have traditionally relied on pure human analysts \citep{guanter2021mapping}, but this is impractical for operational use. Semi-automated methods using statistical thresholding and adjacency criteria have been developed to detect methane enhancements above background levels for plume masking \citep{varon2019satellite, duren2019california, gorrono2023understanding,roger2023high}. However, these methods are susceptible to retrieval artifacts, often mistaking surface features for methane plumes \citep{cusworth2019potential,bruno2023u}.

\subsection{Drawbacks}
Traditional methods for methane plume detection and estimation are significantly hampered by their need for extensive manual intervention. Despite offering reliable enhancements through explainable approaches, these methods suffer from high false detection rates and lack automation, as current techniques often require manual inspection by experts who analyze pre-computed spectral ratio products. This reliance on human analysis makes the process labor-intensive and inefficient. Additionally, these methods are vulnerable to retrieval artifacts, particularly from surface features mistakenly identified as methane plumes, further exacerbating false detection issues.

From the perspective of methane concentration retrieval, traditional techniques struggle with accurately distinguishing true methane enhancements from background noise and artifacts. The manual processes involved in identifying and masking plumes are time-consuming and prone to inconsistencies and errors, leading to unreliable concentration estimates \citep{ruuvzivcka2023semantic}. Moreover, the variability in wind speed characterization complicates accurately estimating methane point source rates. Effective wind speed must account for vertical averages over the plume extent, yet traditional methods often lack the capability to capture detailed horizontal wind variability across the plume's scale. This limitation results in less precise source rate estimates, as the methods cannot consistently account for the complex wind dynamics influencing plume dispersion.

Thus, despite advancements in methane detection techniques, traditional methods remain constrained by their dependence on manual intervention. They are susceptible to inaccuracies in methane concentration retrieval and point source rate estimation. These limitations highlight the need for more automated and robust approaches that can effectively handle the intricacies of plume detection and quantification without extensive human oversight.

\section{Machine learning approaches for methane applications}
With the increasing popularity of \acrshort{ml} in the last decade, scientists have recently employed modern \acrshort{ml} methods for methane detection and quantification of emission rates. \acrshort{ml} approaches, first applied to general computer vision tasks (gain understanding from digital images), are widely applied to the methane domain. As explained in Chapter \ref{chap:traditional_approaches}, the classical methods rely on complex and computationally intensive simulations to produce consistent and accurate results. For \acrshort{ml} models, less domain expertise and no simulation are needed, which results in higher automation, faster processing, and energy-efficient methane detection. For example, \cite{schuit2023automated} employed \acrshort{ml} models for faster methane detection for \acrshort{s5p} at the \acrfull{sron}, reducing the overall processing time. In this Chapter, we will review and summarize the \acrshort{ml} approaches used for methane plume segmentation and emission rate estimation tasks. In addition, some researchers (e.g. \cite{radman2023s2metnet, si2024unlocking}) have developed methods to estimate the emission rate directly from satellite observations.
 Some of the models also include wind information, which may substantially increase the uncertainty of the prediction.

The most used \acrshort{ml} architecture hereafter is based on \acrshort{cnn} and transformer. The fundamental idea for \acrshort{cnn} is to use learnable kernels, which will be convolved with the input data in a 2D image-like form. This operation gives the model great flexibility to focus on different properties in a local neighborhood of each pixel \citep{ronneberger2015u, gu2018recent}. Transformers combine multiple attention layers and \acrfull{mlp} to learn relationships between the queries (Q) and the keys (K) \citep{vaswani2017attention}, resulting in the relationship between all the different pixels. Transformers were applied to text analysis first and to images shortly after \citep{dosovitskiy2020image}. 

\subsection{Machine Learning methods for plume segmentation}
\label{chap:plume_seg}

Creating a methane plume mask involves a fundamental computer vision task known as image segmentation, where the objective is to partition an image into distinct regions corresponding to different classes, in this case, 'plume' and 'no-plume'. Modern approaches in computer vision rely on state-of-the-art models that incorporate attention mechanisms, which form the basis for transformer models. 
Historically, image segmentation models transitioned from traditional \acrshort{ml} classifiers like decision trees and support vector machines to deep learning architectures. Deep learning models such as perceptrons, \acrshort{mlp}, \acrshort{cnn}, and transformers excel at feature extraction and reasoning, making them well-suited for tasks like image segmentation. 
Despite the growing application of \acrshort{ml} in various remote sensing domains, its utilization remains limited due to challenges such as insufficient annotated training data and the complexities inherent in unsupervised or semi-supervised methods. Still, there has been a recent uptick in adopting \acrshort{ml} techniques for methane plume segmentation. However, the volume of research in this area still lags behind other domains within remote sensing, like cloud detection. Cloud detection is a related segmentation task using remote sensing data in which different \acrshort{ml} approaches are frequently employed.

This chapter will provide an overview of used datasets and metrics before analyzing different \acrshort{ml} approaches.

\subsubsection{Datasets for plume segmentation}

Many authors produced data for training plume segmentation models by simulating methane plumes and superimposing them into real observations. Some authors put plenty of effort into the manual annotation of the plumes. To provide an overview of which datasets for which mission exist and can be accessed, we consolidated the datasets with some information in Table \ref{tab:dataset_seg}. As shown in this Table, the amount of data is still relatively sparse for deep learning methods, with just three openly available datasets. \acrshort{avirisng} data has been widely used as this mission provides the most accessible annotated data. Some more datasets could be used for plume segmentation, like \cite{DVN/KRNPEH_2021} providing simulated Sentinel 2 data, but these datasets do not provide a plume mask for ground truth. The \acrshort{enmap} dataset \citep{si2024unlocking} consists of all available \acrshort{enmap} bands that are not sensitive to water vapor or \acrshort{co2}, resulting in 41 bands. 
In Table \ref{tab:dataset_seg}, the \textit{Mission} describes the mission names or the instrument in the case of \acrshort{avirisng}. The \textit{Bands} describe the used bands in the dataset and processing level, sometimes interpolated to a consistent spatial resolution marked in parenthesis. The number of samples is given in the \textit{num samples} column, and the \textit{image size} is provided in the number of pixels. The column \textit{Aug} mentions if this dataset is augmented using techniques like rotation or changes in brightness. The \textit{Access} is grouped into per request (\textit{req}), \textit{open} for free access, and ``\textbf{-}" for unknown data access. Whether or not the data has been simulated is stated in \textit{Sim}, and \textit{Ref} describes the reference for further information about the datasets.

\begin{table}[h]
    \caption{Available datasets for Plume segmentation}    
    \centering
    \resizebox{\textwidth}{!}{
    \begin{tabular}{llllllll}
            \hline
            Mission & Bands & \multicolumn{1}{l}{\begin{tabular}[l]{@{}l@{}} Num \\ samples \end{tabular}} & Img size & Aug & Access & Sim & Ref \\ 
            \hline
             \acrshort{s5p} & 1 (L2) & $>$1.800 & 16 x 22 & no & req & no & \cite{lauvaux2022global} \\
            Sentinel 2 & all (10m, L1) & 10.046 & 200 x 200 & no & req & no & \cite{vaughan2024ch4net} \\ 
            Sentinel 2 & all (20m, L1) & 1.650.000 & 128 x 128 & no & - & yes & \cite{rouet2024automatic} \\ 
            \acrshort{avirisng} & \multicolumn{1}{l}{\begin{tabular}[l]{@{}l@{}}RGB and\\ enhancement\end{tabular}}  & 167.825 & 512 x 512 & yes & open & no & \cite{ruuvzivcka2023semantic}             \\ 
            \acrshort{avirisng} & \multicolumn{1}{l}{\begin{tabular}[l]{@{}l@{}}\acrshort{rgb} and\\ enhancement\end{tabular}} & 46 & 23k x 1.5k & no & open & no & \cite{thompson2017geo} \\
            \acrshort{avirisng} & all (L1) & 3.961 & 256 x 256  & no & - & no & \cite{kumar2023methanemapper} \\
            \acrshort{avirisng} & all (L1) & 7.000 & - & no & - & yes & \cite{jongaramrungruang2022methanet} \\
            GHGSat-C1 & 1 (L2) & 6.870 & 128 x 128  & yes & open & yes & \cite{bruno2023u} \\
            \acrshort{enmap} & \multicolumn{1}{l}{\begin{tabular}[l]{@{}l@{}}41 (L1) and\\ enhancement\end{tabular}} & 7200 & 256 x 256 & no & - & yes & \cite{si2024unlocking} \\ 
            \hline
            \end{tabular}
    } 
    \label{tab:dataset_seg} 
\end{table}

\subsubsection{Evaluation metrics}
The authors using \acrshort{ml} for methane plume segmentation employ a variety of metrics, which complicates model comparison. The commonly used metrics are described below \citep{rainio2024evaluation}.

\begin{equation}
\begin{split}
\mathrm{Accuracy} = \frac{TP + TN}{TP + TN + FP + FN};   ~~~~    \mathrm{Precision} = \frac{TP}{TP+FP}; \\ 
\mathrm{Recall} = \frac{TP}{TP+FN};~~~~ \mathrm{IoU} = \frac{TP}{TP + FP + FN}; ~~~~\mathrm{F1} = 2* \frac{\mathrm{Precision} * \mathrm{Recall}}{\mathrm{Precision} + \mathrm{Recall}}
\end{split}
\end{equation}

\begin{equation}
FPR = \frac{FP}{FP + TN};   ~~~~    FNR = \frac{FN}{FN + TP}; 
\end{equation}

The used metrics for methane plume segmentation, similar to those employed in cloud detection, are Precision \citep{schuit2023automated, rouet2023autonomous, vaughan2024ch4net, groshenry2022detecting, kumar2020deep}, F1-Score \citep{schuit2023automated, ruuvzivcka2023semantic, groshenry2022detecting, kumar2020deep, rouet2024automatic}, Recall \citep{schuit2023automated, vaughan2024ch4net, groshenry2022detecting, kumar2020deep}, Accuracy \citep{schuit2023automated, jongaramrungruang2022methanet, vaughan2024ch4net}, \acrfull{iou} \citep{bruno2023u, groshenry2022detecting, kumar2020deep}, \acrfull{fpr} \citep{rouet2023autonomous, vaughan2024ch4net, ruuvzivcka2023semantic}, \acrfull{miou} \citep{groshenry2022detecting, kumar2023methanemapper}. Single publications also utilized \acrfull{map} \citep{kumar2023methanemapper}, \acrfull{ap} \citep{si2024unlocking} \acrfull{fnr} \citep{vaughan2024ch4net}, \acrfull{tpr} \citep{rouet2023autonomous}, and \acrfull{auprc} \citep{ruuvzivcka2023semantic}. However, \cite{joyce2023using} did not provide any metric for plume segmentation performance.

While some metrics are available for comparison, difficulties arise when authors present performance graphically for each sample instead of providing averages. Additionally, some authors adjust metrics based on \acrshort{snr}, considering different scenes, while others assume a homogeneous scene and assess algorithms based on emission rate. Variations in surface characteristics lead to fluctuations in instrument-measured radiance, impacting estimated methane concentrations and emission rates. 
For the comparability of the various presented models, objective evaluations on standardized benchmark datasets would be needed. A robust comparison of \acrshort{ml} models requires extensive datasets with comparable statistics regarding emission rate, albedo, viewing angle, radiance, scenes, wind speed, and other parameters for each sensor. Such datasets would facilitate clear comparisons and enable evaluations across different sensors. Moreover, these datasets could serve as benchmarks for new methods, offering insights into overall performance and comparisons with existing techniques.

\subsubsection{Approaches for plume segmentation}
We will shortly discuss \acrshort{ml} approaches for cloud detection to show a variety of different approaches that could be applied to methane plume segmentation.

For cloud detection, \cite{foga2017cloud} and \cite{qiu2019fmask} utilized decision tree models like See5 and Automated Cloud Cover Assessment, yielding commendable performance, albeit with lower accuracy compared to more recent methods. Singular value decomposition was employed by \cite{xie2017multilevel} to generate pertinent features for \acrfull{svm} while exploring clustering methods such as K-means for grouping satellite images. Probabilistic approaches applied to handcrafted features using a Bayesian classifier were studied by \cite{mahajan2020cloud}, who also compared the results with random forest, fuzzy logic approaches, boosted random forests, and probabilistic latent semantic analysis. However, these methods exhibited inferior performance compared to deep learning and \acrshort{cnn}.

Especially the \acrshort{cnn}-based U-net architecture \citep{ronneberger2015u} was widely applied. 
Enhancements to the U-net model have been proposed by \cite{li2023cloud}, who initially utilized two U-net structures for segmentation, identifying possible cloudy regions and refining the selection through the second U-net. Combination approaches involving \acrshort{cnn} and \acrshort{mlp}, ensemble methods employing voting mechanisms for pixel classification, and the use of alternative architectures like \acrshort{rs}-Net, \acrshort{vgg-16}, SegNet, or Pyramid Scene Parsing Network have also been explored \cite{li2021review}.

Furthermore, strategies such as superpixel grouping through Simple Linear Iterative Clustering followed by \acrshort{cnn} processing for cloud detection were introduced by \cite{xie2017multilevel}. Inspired by transformer attention modules, \cite{li2023cloud} integrated attention modules into the skip-connections of the U-net architecture. Additionally, the Vision Transformer was employed for cloud detection by \cite{zhang2023measuring}. Recent semi-supervised approaches have introduced the Swin Transformer, combined with \acrshort{cnn} and multi-head attention mechanisms for cloud detection by \cite{gong2023hybrid}.

\begin{figure}
    \centering
    \includegraphics[width=1\textwidth]{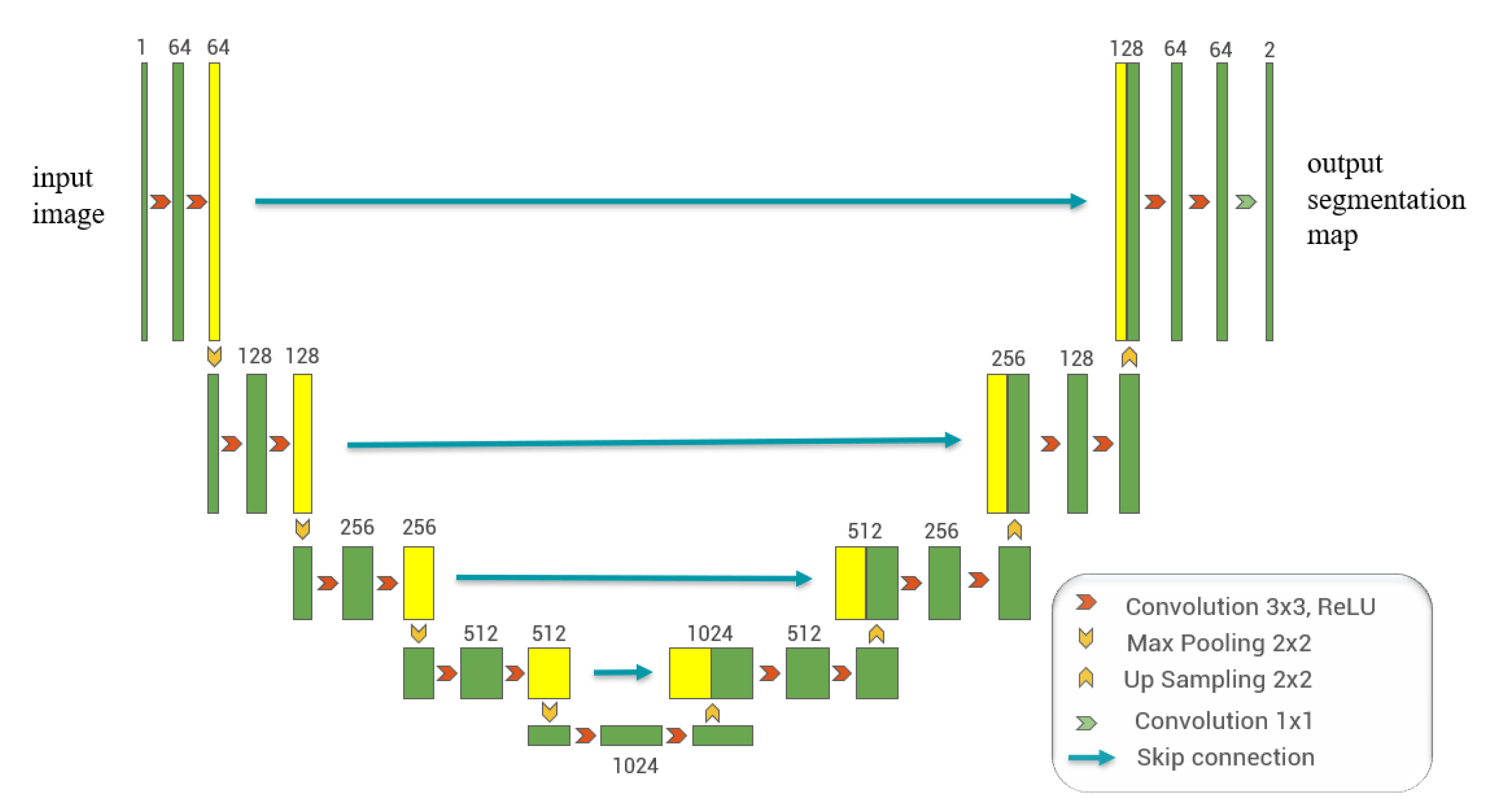}
    \caption{This image by \cite{guo2020cloud} depicts the U-net architecture of \cite{ronneberger2015u} under a CC-BY license. The U-net architecture has been used in different forms for recent methane plume segmentation approaches. The architecture uses different convolution sizes, reduces the image size first (encoder) to reduce the feature space, and decodes the image (decoder) back to its original shape, additionally providing skip connections for more effective learning.}
    \label{fig:architecutre_clac_U-net}
\end{figure}

Various methodologies have been explored in the cloud detection domain, contrasting with methane plume segmentation where \acrshort{cnn} models predominate, especially based on the U-net architecture shown in Figure \ref{fig:architecutre_clac_U-net}. This preference for \acrshort{cnn} may stem from their superior performance and ability to retain spatial information, a quality often lost in classifiers like decision trees or \acrshort{svm} \citep{schuit2023automated}. 
Next, we would like to discuss the applied \acrshort{ml} approaches for methane plume segmentation. 

\begin{figure}
    \centering
    \includegraphics[width=1\textwidth]{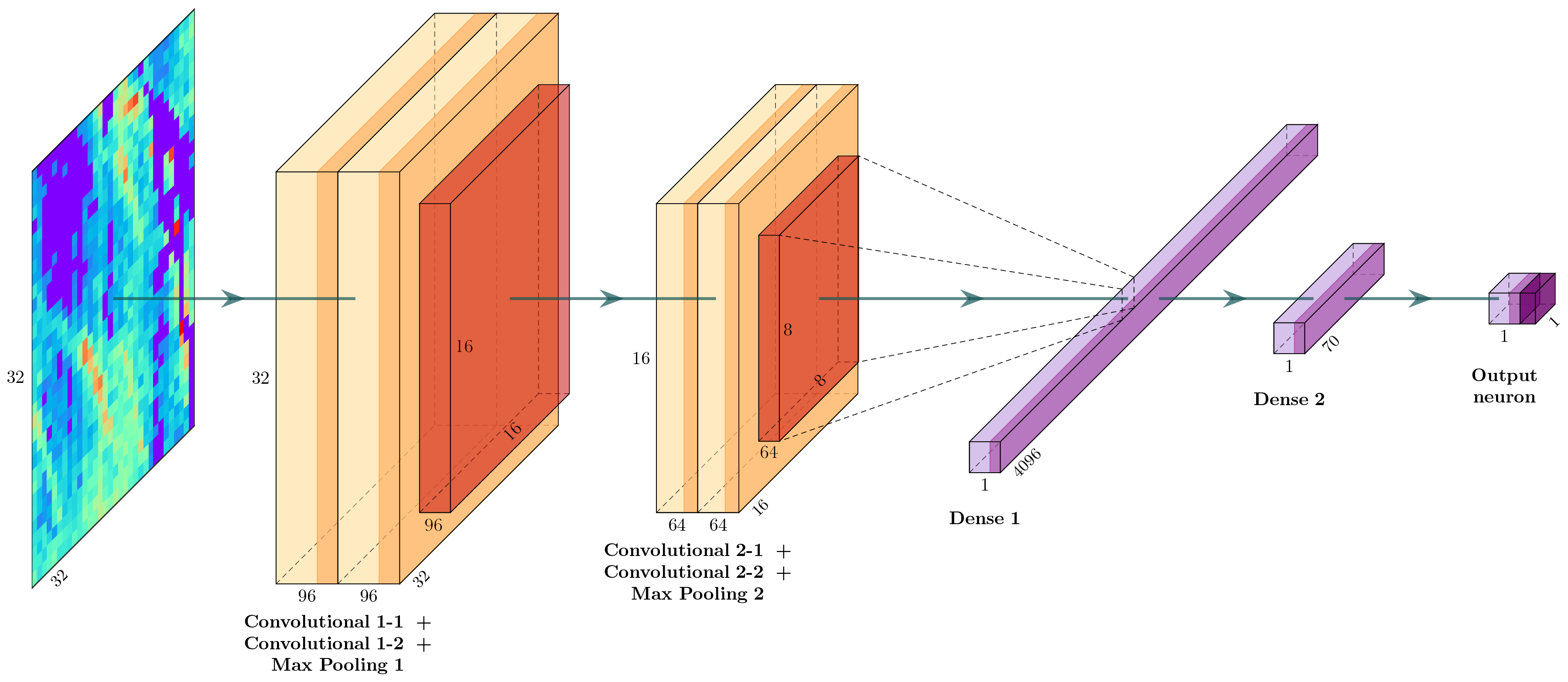}
    \caption{Architecture used by \cite{schuit2023automated} under the CC-BY 4.0 license}
    \label{fig:architecutre_Schuit}
\end{figure}

One smaller \acrshort{ml} architecture has been used by \cite{schuit2023automated}, which employed a \acrshort{cnn} encoder (consisting of two \acrshort{cnn} layers) with a classification head (comprising two \acrshort{mlp} layers) for methane plume segmentation using pre-processed  \acrshort{s5p} observations. The architecture is shown in Figure \ref{fig:architecutre_Schuit}. Their study revealed that deeper networks did not yield improved performance. The authors introduced a plume shape and class activation map \citep{zhou2016learning}, along with hand-selected features, into a Support Vector Classifier to discern between real plumes and artifacts through binary classification. Similarly, \cite{finch2022automated} employed a similar architecture for \acrfull{nox} plume segmentation in \acrshort{s5p} data.
A slightly larger model has been employed by \cite{jongaramrungruang2022methanet}, who developed Methanet primarily for emission rate estimation but adapted the model's head to a binary classification of plume presence using 4 convolutional layers with 2 \acrshort{mlp} layer trained on simulated \acrshort{avirisng} data with an accuracy of 90\% for emission rate $>$100 kg/h and 50\% for emission rate of 50-60 kg/h.

The next works employ the standard or modified version of the U-net model. The first use of the U-net architecture was by \cite{groshenry2022detecting}, which generated transformed plumes and methane concentration maps from Sentinel 2 to \acrshort{prisma} scenes for training a U-net model. The authors reported that the model trained solely on \acrshort{prisma} data performed better than the training on Sentinel 2 plumes adapted to the \acrshort{prisma} images. However, the comparison and transformation of Sentinel 2 plumes into \acrshort{prisma} scenes is not clearly described. Following the same architecture \cite{bruno2023u} focused on the widely used U-net architecture for GHGSat image segmentation, training the model on simulated plumes superimposed onto GHGSat plume-free observations. They evaluated performance using the Jaccard score (Intersection over Union), resulting in improved results in low background noise scenarios. CH4Net has been introduced by \cite{vaughan2024ch4net}, training a 4 Block U-net on hand-annotated Sentinel 2 observations, achieving enhanced performance by integrating all available bands into the model. The authors interpolated the bands to 10m resolution and found improved performance when integrating all bands. Utilizing simulated Gaussian plumes superimposed onto Sentinel 2 images for training, a U-net model has been employed by \cite{rouet2023autonomous}. They employed all band ratios from two consecutive Sentinel 2 images, interpolated to 20 m resolution, and achieved improved results compared to the \acrshort{mbmp} \citep{varon2021high} (using a threshold) method for low \acrfull{snr}. Another model for \acrshort{prisma} data training a U-net on simulated plumes superimposed onto real observations, incorporating an additional 1x1 convolutional layer with 64 filters for robustness towards anomalies and an auxiliary loss function to address plume class imbalance, has been trained by \cite{joyce2023using}. 

The U-net architecture and a more sophisticated computer vision model have been used by \cite{ruuvzivcka2023semantic}. The authors utilized \acrshort{avirisng} observations with manually annotated plumes and simulated Worldview-3 images for training two different U-net architectures, employing the MobileNet-v2 \citep{sandler2018mobilenetv2} encoder and the standard U-net decoder for each sensor. The authors provided the methane enhancement map and \acrshort{rgb} images of the scene as input for the model, resulting in lower detection of artifacts and, therefore less false positives. More complex \acrshort{cnn} models have been used by \cite{kumar2020deep}, who used a new approach within plume segmentation by using the ensemble method. The authors used all \acrshort{avirisng} bands and grouped them for input into matched filters for multiple methane maps. These maps are forwarded into a Mask-RCNN \citep{he2017mask}, each using the Resnet101 \citep{he2016deep} for the backbone model. The backbone has been pre-trained on the \acrfull{coco} dataset \citep{lin2014microsoft} and fine-tuned for hyperspectral remote sensing. The output of this ensemble is then fused into an ensemble network (2-layer \acrshort{mlp} also using terrain information) for plume segmentation. Within a second work, the Mask-RCNN has been employed by \cite{si2024unlocking} using \acrshort{prisma} and \acrshort{enmap} data using simulated plumes. The authors compared the use of raw hyperspectral data with concentration maps and a multi-task model combining two models (Mask R-CNN for plume segmentation and a ResNet-50 for emission rate estimation) trained together. They found increased results when using methane concentration maps and the best results when combining the two models. 

\begin{figure}
    \centering
    \includegraphics[width=1\textwidth]{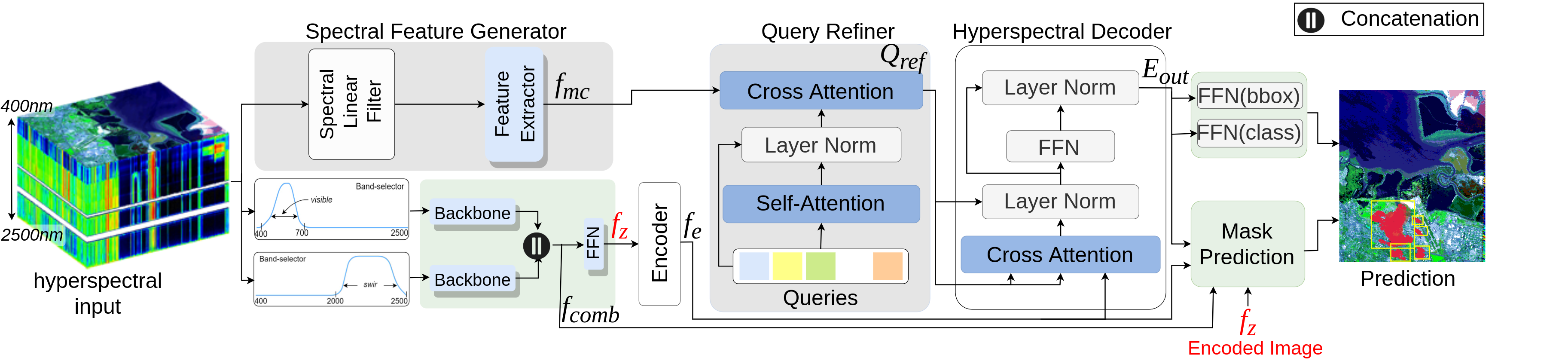}
    \caption{Architecture based on \acrshort{cnn} and transformer elements used by \cite{kumar2023methanemapper} for plume detection and segmentation under the CC-BY 4.0 License. $f_{mc}$ describes potential \acrshort{ch4} feature maps, $f_{comb}$ is the combined output of the Backbones using different spectral ranges, $f_Z$ is the output of two \acrshort{mlp} from $f_{comb}$, $f_e$ is the feature encoded $f_Z$. }
    \label{fig:architecture_kumar2023}
\end{figure}

So far, all authors have used \acrshort{cnn} as a basis model for plume segmentation. However, some authors were able to apply the first transformer model for plume segmentation/detection. 
\cite{kumar2023methanemapper} used a pre-trained detection transformer and pre-trained ResNet \citep{he2016deep} model together using \acrshort{avirisng} observation for plume segmentation and bounding box around the plume. The authors use a spectral feature generator (selector and \acrshort{cnn} model backbone) to generate methane plume candidates to be used in a query refiner. Parallel to that, methane-sensitive bands use the \acrshort{cnn} backbone for feature generation, which will be embedded for cross-attention between the embedding and the query refiner for the decoder transformer model. The output is used for a pyramid model (2 layer \acrshort{mlp}) to generate the bounding box and a mask predictor using the methane-sensitive features, encoded image, and encoded features. The mask predictor consists of the segmentation head of the \acrfull{detr} \citep{carion2020end}. Heat maps for each plume will be produced and filtered by a threshold for plume mask generation. The complex architecture is depicted in Figure \ref{fig:architecture_kumar2023}.

The recent work of \cite{rouet2024automatic} built a large dataset using simulated Gaussian plumes superimposed into Sentinel 2 images using the Beer-Lambert law \citep{ehret2022global} from different countries across the world. The authors then trained a model using a vision transformer encoder and \acrshort{cnn} decoder from scratch, following a U-net architecture for pixel-wise binary classification. The trained model provides a low false positive rate, better detection at low \acrshort{snr} compared to \acrshort{mbmp} with threshold, and significantly lowered the detection limit of Sentinel 2 down to 200-300 kg/h. The detection limit has been tested on further detected plumes from \acrshort{avirisng} and \acrfull{gao}, labeling a plume as detected if at least 2 consecutive pixels are marked as 'plume', which is not more than 500 m away from the beforehand detected source location. However, it is unclear if the detected plume pixels are valid pixels of a plume and need further investigation.

In plume segmentation using \acrshort{ml}, the prevailing approach typically involves simple or U-net structured \acrshort{cnn}, yielding higher accuracy than threshold-based methods. 
Due to the labor-intensive nature of generating ground truth plume masks from real observations, many authors resort to simulated data tailored to the specific sensor. To approximate real-world conditions, simulated plumes are often overlaid onto actual observations devoid of plumes, capturing realistic background and sensor noise. In contrast to cloud detection, probabilistic approaches for plume segmentation were not prevalent at the time of writing. Performance metrics for methane plume segmentation vary, posing challenges in comparing different methods. Another hurdle is comparing different satellite sensors, such as \acrshort{avirisng} and Sentinel 2, which exhibit discrepancies in spectral and spatial resolutions and surface characteristics, leading to variations in \acrshort{snr} for identical plumes. Even among studies utilizing the same sensor, such as \acrshort{avirisng} data \citep{jongaramrungruang2022methanet, ruuvzivcka2023semantic, kumar2020deep, kumar2023methanemapper}, comparability is hindered by the adoption of different metrics, with some studies altering metrics mid-research \citep{kumar2020deep, kumar2023methanemapper}, further complicating cross-study comparisons.

\subsection{Machine Learning methods for emission rate estimation}

The task of estimating the emission rate of a particular methane plume is, from an \acrshort{ml} perspective, a general image-level regression task. 
Regression describes the challenge of finding correlations of the given inputs, which could be independent or dependent variables on the desired output. It is commonly the task to minimize the distance between the desired output and the model's prediction. 
Next, we will describe the available datasets, metrics, and different \acrshort{ml} approaches for emission rate estimation.

\subsubsection{Data sets for emission rate estimation}

The \acrshort{ml} approaches for emission rate estimation rely on supervised training and, therefore, on many ground truth data. For this reason, many authors relied on simulated plumes superimposed on the specific observation. Table \ref{tab:dataset_reg} lists the used datasets for the \acrshort{ml} approaches and further available datasets. In addition to the Table, different authors used plume simulations to find a suitable effective wind speed for the \acrshort{ime} method or for validating sensors. However, these datasets are limited to specific areas and contain a small amount of samples in most cases, but the generated synthetic dataset is often not further described or released. For example, this has been done, e.g., by \cite{roger2023exploiting}. 

In Table \ref{tab:dataset_reg}, the \textit{Mission} is named or the instrument in the case of \acrshort{avirisng}. The \textit{Bands} represent the number of used bands in the dataset and processing level. If all available channels of the satellite are provided, the number of bands is set to ``all". The number of samples and different scenes is given in the \textit{samples} column, and the \textit{image size} is provided in the number of pixels. The column \textit{Aug} mentions if this dataset is augmented using techniques like rotation or changes in brightness. Next, the \textit{minimum and maximum used emissions} from the simulation in kg/h is described. Accounting for the different wind speeds, the \textit{Wind} column provides the modeled speeds. The \textit{Access} is grouped into per request ``req'', open for free access and no for no public access. \textit{Ref} describes the reference for further information about the datasets and ``\textbf{-}" describes missing values.

\begin{table}[h]
    \caption{Available datasets for emission rate estimation providing simulated ground truth values} 
    \centering
    \resizebox{\textwidth}{!}{
    \begin{tabular}{lllllllll} 
            \hline
            Mission & Bands & \multicolumn{1}{l}{\begin{tabular}[l]{@{}l@{}} Samples \\ (Scenes) \end{tabular}} & Img size & Aug &  \multicolumn{1}{l}{\begin{tabular}[l]{@{}l@{}} Min / \\ Max \\ emission \\ (kg/h) \end{tabular}} & \multicolumn{1}{l}{\begin{tabular}[l]{@{}l@{}} Wind \\ m/s  \end{tabular}} & Access & Ref \\   
            \hline
            Sentinel 2 & 1 (L2) & \multicolumn{1}{l}{\begin{tabular}[l]{@{}l@{}} 7000 \\ 3 \end{tabular}} & \multicolumn{1}{l}{\begin{tabular}[l]{@{}l@{}} 100x100 \\ \end{tabular}} & no & \multicolumn{1}{l}{\begin{tabular}[l]{@{}l@{}} 5k \\ 30k \end{tabular}} & 1-10 & req & \cite{radman2023s2metnet} \\ 
            Sentinel 2 & all (L1) & \multicolumn{1}{l}{\begin{tabular}[l]{@{}l@{}} 1345 \\ 3 \end{tabular}} & \multicolumn{1}{l}{\begin{tabular}[l]{@{}l@{}} 75x75 \\ (to 20m) \end{tabular}} & no & \multicolumn{1}{l}{\begin{tabular}[l]{@{}l@{}} 500 \\ 50k \end{tabular}} & 3.5 & open & \cite{DVN/KRNPEH_2021} \\ 
            \acrshort{prisma} & all (L1) & \multicolumn{1}{l}{\begin{tabular}[l]{@{}l@{}} 9700 \\ 36 \end{tabular}} & \multicolumn{1}{l}{\begin{tabular}[l]{@{}l@{}} 256x256 \\  \end{tabular}} & no & \multicolumn{1}{l}{\begin{tabular}[l]{@{}l@{}} 100 \\ 10k \end{tabular}} & 1-9 & open & \cite{joyce2023using} \\ 

            GHGSat-C1 & 1 (L2) & \multicolumn{1}{l}{\begin{tabular}[l]{@{}l@{}} 6870 \\ 28 \end{tabular}} & \multicolumn{1}{l}{\begin{tabular}[l]{@{}l@{}} 128x128 \\  \end{tabular}} & yes & \multicolumn{1}{l}{\begin{tabular}[l]{@{}l@{}} 500 \\ 2k \end{tabular}} & 3-9 & open & \cite{bruno2023u} \\ 

            \acrshort{avirisng} & all (L1) & \multicolumn{1}{l}{\begin{tabular}[l]{@{}l@{}} 7000 \\ - \end{tabular}} & \multicolumn{1}{l}{\begin{tabular}[l]{@{}l@{}} 300x300 \\  \end{tabular}} & no & \multicolumn{1}{l}{\begin{tabular}[l]{@{}l@{}} 0 \\ 2k \end{tabular}} & 0-12 & no & \cite{jongaramrungruang2022methanet} \\ 

            WorldView3 & all (L1) & \multicolumn{1}{l}{\begin{tabular}[l]{@{}l@{}} - \\ 3 \end{tabular}} & \multicolumn{1}{l}{\begin{tabular}[l]{@{}l@{}} 406x406 \\  \end{tabular}} & no & \multicolumn{1}{l}{\begin{tabular}[l]{@{}l@{}} 100 \\ 3k \end{tabular}} & 3.5 & req & \cite{sanchez2022mapping} \\

            Sentinel 2 & all (L1) & \multicolumn{1}{l}{\begin{tabular}[l]{@{}l@{}} - \\ 700 \end{tabular}} & \multicolumn{1}{l}{\begin{tabular}[l]{@{}l@{}} 406x406 \\ (to 20m) \end{tabular}} & no & \multicolumn{1}{l}{\begin{tabular}[l]{@{}l@{}} 100 \\ 3k \end{tabular}} & 3.5 & req & \cite{rouet2023autonomous} \\ 

            \acrshort{enmap} & 42 (L1) & \multicolumn{1}{l}{\begin{tabular}[l]{@{}l@{}} 7200 \\ 1 \end{tabular}} & 256 x 256 & no & \multicolumn{1}{l}{\begin{tabular}[l]{@{}l@{}} 500 \\ 2000 \end{tabular}} & 1-10 & - & \cite{si2024unlocking} \\ 

            \hline
            \end{tabular}
    \label{tab:dataset_reg} 
  } 
\end{table}

\subsubsection{Evaluation metrics}
For estimating emission rates, methods can only be compared effectively if simulated plumes are utilized to establish a ground truth for evaluation. Consequently, all authors employing \acrshort{ml} approaches for emission rate estimation have relied on simulated plumes. Some authors have conducted small-scale controlled emission experiments to validate existing methods and generate ground truth, particularly for \acrshort{avirisng} or other flight campaigns. Recently, \cite{sherwin2024single} conducted a single-blind evaluation of different satellite sensors. They synchronized controlled emissions with satellite overpasses and measured the emitted gas, resulting in multiple satellite images. Although these images cannot be directly used for \acrshort{ml} or deep learning approaches, as they result in 2 to 5 images from different satellites, they could be used to validate trained models.

The widely used metrics for regression, in general, are the \acrfull{mape}, \acrshort{rmse}, and \acrfull{r} described below \citep{naser2023error}.

The \acrshort{mape} is defined as
\begin{equation}
        MAPE = \frac{100}{n} \sum_{i=1}^{n} \left| \frac{y_i - \hat{y}_i}{y_i} \right| ,
\end{equation}
where $n$ is the number of data points, $y_i$ is the i-th value, $\hat{y}_i$ describes the i-th prediction of the model.

\begin{equation}
    RMSE = \sqrt{\frac{1}{n} \sum_{i=1}^{n} (y_i - \hat{y}_i)^2},
\end{equation}
where $y_i$, $\hat{y}_i$ are the same values as above for the \acrshort{mape}.

The Pearson correlation coefficient is defined as \citep{rainio2024evaluation}:
\begin{equation} 
    R = \frac{\sum_{i=1}^{n} (x_i - \bar{x})(y_i - \bar{y})}{\sqrt{\sum_{i=1}^{n} (x_i - \bar{x})^2 \sum_{i=1}^{n} (y_i - \bar{y})^2}},
\end{equation}
where the data points are describes as $x_i$ and $y_i$ and $\bar{x}$ and $\bar{y}$ are the respective means of the data.

For comparing the efficacy of different models, authors typically rely on \acrshort{mape} \citep{jongaramrungruang2022methanet, radman2023s2metnet, joyce2023using}. \cite{radman2023s2metnet} additionally included multiple metrics such as \acrshort{rmse} and Pearson correlation coefficient. Similar \cite{si2024unlocking} measured the performance by the \acrshort{rmse} and the \acrfull{mae}. However, \cite{bruno2023u} did not provide any performance metrics. Moreover, \cite{bruno2023u} and \cite{joyce2023using} presented performance results in plots, further constraining direct comparison.

Due to differences in datasets, simulations, and atmospheric parameters, the metrics used across studies remain non-comparable, akin to challenges encountered in plume segmentation tasks. Harmonizing datasets, as described in Chapter \ref{chap:plume_seg}, would be essential for facilitating meaningful comparisons in this task.

\subsubsection{Approaches for emission rate estimation}
Within the field of computer vision, a common regression task would be the depth estimation of an image. In this domain, the best-performing models use specialized convolutions, such as fast Fourier convolutions \citep{Berenguel_Baeta_2023} or attentively combined dilated convolutions \citep{zhuang2022acdnet} or are based on transformer architecture or a combination of \acrshort{cnn} and transformer \citep{junayed2022himode}. 
Typical metrics for comparison are \acrfull{rmse} or the \acrfull{are} shown in Equation \ref{eqn:RMSE_ARE_description} \citep{hodson2022root},

\begin{equation}
\label{eqn:RMSE_ARE_description}
    \mathrm{RMSE} = \sqrt{\frac{1}{n} \sum_{i=1}^{n} (y_i - \hat{y}_i)^2} , ~~~ \mathrm{ARE}=\frac{|y_i - \hat{y}_i|}{|y_i|} 
\end{equation}
where $n$ is the number of data points, $y_i$ is the i-th value, $\hat{y}i$ describes the i-th prediction of the model.

For remote sensing regression tasks, finding a domain similar to emission rate estimation adopting various \acrshort{ml} approaches is difficult. One regression domain is to determine soil parameters using \acrshort{ml} approaches \acrfull{mlr}, Random forest regression, \acrshort{svm} for regression or Stochastic gradient boosting \citep{forkuor2017high}. Generally, the more complex \acrshort{ml} approaches show improved performance to \acrshort{mlr} using \acrshort{rmse} \citep{forkuor2017high}. 
Within the global land surface temperature domain, \acrshort{mlp} has been used \citep{cifuentes2020air}. For estimating land surface temperature, \acrshort{cnn}-based models have been used by \cite{tan2019deep} or even 3D-CNN by \cite{fu2022combining}, providing promising results.

Like plume segmentation, the different \acrshort{ml} approaches for emission rate estimation are scarce. The various \acrshort{ml} classifiers or transformers have not been tested for now. Some of the mentioned emission rate methods (see Chapter \ref{chap:review_emission_rate_traditional}) rely on wind speed and direction. 

Wind information is a crucial parameter for the traditional approaches, while some ML approaches do not contain any wind information. 
Wind information for each pixel results from weather forecasts on a much coarser grid most of the time, as accurate wind data is often unavailable. ERA5, the weather prediction model from the \acrfull{ecmwf}, provides worldwide weather data on a 9x9 km grid (using ERA5-Land). In contrast, the \acrfull{gfs} of the \acrfull{nws} provides a 28x28 km resolution. As the grid information is more coarse than the pixel size of most satellites, it results in uncertainty, while the information itself is a prediction that includes uncertainty and may vary for different locations. Therefore, many authors assume an uncertainty in wind information of 50\% \citep{roger2024high, joyce2023using, guanter2021mapping}. 
High uncertainties arise in emission rate estimation due to varying boundary layer conditions like turbulence and wind speed, spectral interferences, and the sensitivity of the flux inversion to complex plume structures. Using traditional approaches, the overall uncertainties can be up to 100\% for individual plumes \citep{duren2019california}. 
 It can be observed that some models without wind data provide a lower uncertainty. Whether or not the inclusion of wind information into the model provides benefits or impacts the uncertainty of the model will be mentioned by the different authors and summarized.

The first deep learning approach for emission rate estimation has been implemented by \cite{jongaramrungruang2022methanet}. The authors tested the same model used for plume segmentation for regression tasks, achieving favorable results across wind speeds and emission rates, even without explicit integration of wind information. They identified an emission rate limit of approximately 50 kg/h for their model, which reduced error ranges from 15-65\% using the traditional \acrshort{ime} method to 21\% without wind information. Instead of using the enhancement maps as the only input, \cite{joyce2023using} employed multiple \acrshort{cnn} models for plume segmentation, methane concentration estimation, and binary classification (presence of plume in the image) as inputs for an emission rate \acrshort{cnn} model. Their model, comprising 8 convolutional layers, achieved a median error of 25\%, compared to approximately 50\% using traditional methods. Utilizing multiple models for intermediate steps resulted in improved performance compared to \cite{jongaramrungruang2022methanet}. Compared to these models \cite{bruno2023u} conducted a comparative study between a \acrshort{cnn}-based model and the \acrshort{ime} method for emission rate estimation using simulated plumes in GHGSat images. While the authors did not explicitly describe the architecture of their model, examination of a provided Jupyter notebook revealed a model comprising 2 convolutional layers, 4 down convolutional layers, and a regression head (comprising 2 \acrshort{mlp} layers and Rectified Linear Units (\cite{DVN/YFRQU4_2023}) which is similar to the architecture in Figure \ref{fig:architecutre_Schuit}. They observed a smoothing bias in their model, which tended to overestimate low emission rates and underestimate large emission rates, similar to findings by \cite{jongaramrungruang2022methanet} and \cite{joyce2023using}. However, the \acrshort{cnn} model exhibited greater robustness due to lower uncertainty. However, wind speed from the simulation has been integrated, providing no error in wind information, whereas wind direction integration had negligible effects. Notably, errors in wind prediction dominated when wind speeds were below 4 m/s, while plume mask errors dominated for higher wind speeds. Emission rate estimation for Sentinel 2 has been conducted by \cite{radman2023s2metnet}, which employed simulated plumes in observations to estimate emission rates using various deep learning methods, including \acrshort{cnn}-based models and a transformer model \citep{he2016deep, szegedy2015going, simonyan2014very, huang2018urban, tan2021efficientnetv2, liu2021swin}. These models, operating without wind information, consistently underestimated larger emission rates, with \acrshort{cnn} approaches showing stronger underestimation. The authors compared models trained with random initialization, pre-trained on ImageNet \citep{deng2009imagenet}, and fine-tuned on this task, demonstrating the efficiency of fine-tuning pre-trained models. They identified \acrshort{cnn} models as the most suitable for emission rate estimation, outperforming \acrshort{ime} and MethaNet of \cite{jongaramrungruang2022methanet}. Additionally, using pre-trained AlexNet and ResNet-50 (fine-tuned to emission rate estimation) results in better estimations than IME \citep{si2024unlocking}. The authors found a better performance of the ResNet-50 model for the single task of emission rate estimation compared to a multi-task model for the plume segmentation and the emission rate estimation.

The works of the aforementioned authors demonstrated improved performance compared to \acrshort{ime} using \acrshort{cnn}. As for plume segmentation, \acrshort{cnn} dominates this task, with few alternative approaches explored. \cite{radman2023s2metnet} compared transformer models to various \acrshort{cnn} models and found superior performance in \acrshort{cnn}. 
However, all \acrshort{ml} methods tended to underestimate large emissions, albeit to varying degrees. This raises whether alternative \acrshort{ml} architectures or approaches could yield improvements. Notably, unlike \acrshort{ime}, current \acrshort{ml} models do not integrate wind information for emission rate estimation, reducing uncertainty. The assumed mean absolute error of the wind speed is $\pm$ 50\% (\cite{roger2024high, joyce2023using, guanter2021mapping}). 
Investigating the impact of integrating wind information into \acrshort{ml} models warrants further research. Additionally, the authors used two different inputs for the \acrshort{ml} models: either only the enhancement maps like \cite{jongaramrungruang2022methanet, radman2023s2metnet} or the enhancement map together with a plume mask used by \cite{joyce2023using, bruno2023u}. As a realistic comparison is impossible, this would be an additional topic to be investigated, which could be compared using an aligned dataset testing both approaches.

\section{Discussion and Outlook}

Recent studies and models utilizing \acrshort{ml} for methane-related tasks have demonstrated significant advantages, including enhanced accuracy, faster execution times, reduced uncertainty, and increased automation. 
Most of these \acrshort{ml}-based works have been published within the past five years. This trend underscores the growing interest and application of \acrshort{ml} in methane-related tasks. Furthermore, the frequent inclusion of the term “automatic” in the titles of these works indicates a strong desire to automate methane detection processes.

Current \acrshort{ml} models have successfully reduced human intervention for designing specific features, adjusting thresholds, or conducting radiative transfer simulations, promoting a more automated approach. However, manual checks are still required to verify whether the detected plume is an artifact or a genuine plume and whether the plume segmentation was successful. 

One potential solution to this challenge could be the application of explainable AI, which would provide insights into the model’s decisions and foster greater trust in its performance and outcomes. 
This could be used to investigate which parameters or areas the model focuses on and to fine-tune the focus for better prediction. This could lower the necessary manual investigation. 

Combining different models using ensemble methods could further minimize false detections and reduce human involvement. 
The methodology of \cite{pandey2023daily} proposes a rapid and global methane detection system, employing an array of satellites to precisely locate emission sources. This system utilizes satellites with high temporal resolution but coarse spatial resolution and progressively refines the location using satellites with finer spatial resolution. While this system is an excellent foundation, it requires manual intervention at each stage. Integrating \acrshort{ml} and data fusion techniques could automate this process, accelerating the development of a continuous monitoring system for large-scale emissions. However, the high detection limits of satellites with coarse spatial resolution present a constraint that needs to be addressed. 

Looking ahead, we expect the development of more models, potentially featuring complex architectures inspired by applications in the computer vision domain. However, these complex models usually require more training data due to their increased number of trainable parameters. The variety of models could show which kind of model works best in specific domains when providing benchmark datasets. 
This shows the downside of \acrshort{ml} needing large amounts of training data to train complex models and to compare models. In parallel to the development of new models, we also expect a stronger focus on simulated and curated datasets for methane detection. Similar methodology from benchmark datasets within computer vision could be applied to methane tasks for comparison. 

The uncertainty of \acrshort{ml} models and the influence of input parameters (e.g. wind speed) is still unclear and needs in-depth analysis. This analysis might research further benefits and limitations of \acrshort{ml} approaches with respect to specific domains, areas, or situations. The investigation of uncertainty could distinguish downsides for specific use cases, allowing for the selection of appropriate models for particular scenarios that are unknown so far.

The models mentioned here are purely \acrshort{ml} or physics based. Physics-informed or hybrid \acrshort{ml} models combining physics and \acrshort{ml} could improve the overall prediction performance. These models may combine information from image analysis and physical approaches like \acrshort{ime} for a consistent prediction.

For plume segmentation, existing \acrshort{ml} methodologies offer substantial advantages over traditional threshold methods, an advantage that may further increase in the future. \acrshort{ml} techniques excel in distinguishing between plumes and background noise, an area that could see further enhancement through the application of diffusion models. These models can potentially learn and reduce the noise in real-world data, thereby improving plume segmentation. 

Estimating emission rates poses a significant challenge due to the scarcity of ground truth samples and the potential disparity between simulations and real-world data. While more controlled-release experiments would yield additional data, they also pose environmental risks due to the artificial emissions they produce. Given the limited data, these controlled emissions could be more effectively used for model validation rather than training. In this context, further investigation is needed to determine whether compute-intensive physical simulations are necessary to maintain physical consistency within the simulations or if the Gaussian plumes used by \cite{rouet2024automatic} are sufficient for training, given their rapid and voluminous generation. Another potential approach involves creating a surrogate model of the physical plume simulation, possibly through a physically informed \acrshort{ml} model, Generative Adversarial Networks, or generative diffusion models. This would enable the accurate and rapid generation of plumes to create large datasets. The same simulated plume parameters could also be utilized across different datasets, considering different spatial resolutions as other satellite specifications are incorporated at a later stage. Using the same plumes for different sensors could provide a comparative advantage.

The recent release of a legislative act by the EU Parliament aimed at reducing methane emissions underscores the progress being made in combating these emissions \citep{methane_act2021}. This Methane Act mandates that oil and gas producers monitor their assets and report emissions, fostering fast detection and mitigation. This legislation could potentially boost the use of additional satellites for methane detection, thereby enhancing prediction accuracy, expanding observational coverage, and increasing the frequency of monitoring. 
Since most satellites have an expected lifespan of approximately seven years (as per the average in Table \ref{tab:all_sensors}), collecting large real-world datasets of methane plumes globally might not be feasible. However, satellite-specific data simulation could develop a dataset before the satellite’s launch, allowing for extended testing on real-world data.

This work focused on the application of methane, even though the tasks and challenges are similar to other Molecules like \acrshort{co2} or \acrshort{nox}. Further investigation could be done to analyse the synergies between these domains. A potential benefit of cross-domain approaches may be the training of a segmentation model to general plume structure (e.g. \acrshort{co2} plumes) and fine-tune the model for methane segmentation. However, there could be some specific differences that are hard to overcome for a cross-domain model.

\section{Conclusion}

In this study, we compiled existing and future systems for methane detection and compared traditional physics-based approaches with \acrshort{ml} techniques. We explored various multi- and hyperspectral satellite systems capable of methane detection, ranging from coarse resolution with global coverage to fine resolution for local areas. The recent launch of MethaneSat focuses on an intermediate step between coarse area flux mappers and point source mappers. Trends indicate a shift toward specialized greenhouse gas instruments using a few bands or generalized instruments with hyperspectral capabilities. Future systems will focus on high spectral resolution instead of lower GSD, aiming to reduce the detection threshold for even smaller methane plumes. 
The most common methods for methane plume segmentation and emission rate quantification remain traditional physical approaches. Among these, the band ratio technique for multispectral instruments or the matched filter approach for hyperspectral instruments provides an effective and reliable method for methane enhancement calculations. Plume segmentation predominantly relies on thresholding approaches, with innovative methods proposed by researchers such as using a connection of low and high thresholds like by \cite{watine2023geostationary} or connected amount of plume pixel by \cite{schuit2023automated}. 
The \acrshort{ime} method for estimating emission rates considers critical physical parameters such as dissipation and diffusion. It offers an efficient and easily computable algorithm. However, to effectively utilize \acrshort{ime}, one must perform various simulations to approximate the important factor of effective wind speed ($U_{eff}$).

\acrshort{ml} has demonstrated advantages in both detection and emission rate estimation tasks. Existing research indicates improved performance for multiple satellites and airborne sensors using multi- and hyperspectral imagery. These models heavily rely on \acrshort{cnn} and architectures like the U-net. The work of \cite{rouet2024automatic} opens a new path stating that the detection threshold can be drastically improved with a transformer-\acrshort{cnn}-based model and a large training dataset. However, it is unclear if the performance can be generalized. 
Interestingly, some authors observed a smoothing effect in emission rate estimation in the validation dataset, leading to overestimating low and underestimating high emission rates. This smoothing effect results, among other effects, from strong weights of outliers/extremes of the \acrshort{rmse} function \citep{naser2023error}. 
Addressing this issue requires different loss functions or data and a thorough comparison of different deep-learning approaches, which remains relatively unexplored for methane detection. 

Recent efforts have been made to create \acrshort{ml}-ready datasets for \acrshort{avirisng} and Sentinel 2 to facilitate model comparison. However, the \acrshort{avirisng} dataset has limitations as it provides the methane enhancement map and \acrshort{rgb} channels for each sample rather than the hyperspectral data used by other studies (e.g., \cite{kumar2023methanemapper, joyce2023using}). While the simulated Sentinel 2 dataset by \cite{radman2023s2metnet} could serve as a baseline, it is not openly available. Additionally, other available datasets offer limited training data and a small variety of scenes and backgrounds, constraining their use for training deep learning models. 
The main challenge for \acrshort{ml} models remains the scarcity of training samples, especially when dealing with satellite data for estimating emission rates. Generating accurate real-world ground truth data is both expensive and difficult to manage. Ground truth data can either be estimated and compared using different algorithms, as demonstrated in the work by \cite{ruuvzivcka2023semantic}, or based on simulations, as explored in the research of \cite{radman2023s2metnet}. However, models trained on simulated data need further investigation on larger datasets, as many studies have only tested them on a limited number of samples. One potential solution is to provide precise ground truth through simulations and supplement it with a large real-world test set to evaluate model performance on actual data. 

Models should be comparable across different sensors using common objective performance metrics. We anticipate increased adoption of \acrshort{ml} in this domain, resulting in diverse models, exploration of different architectures, and potentially task and domain-agnostic approaches. Additionally, there is a growing interest in foundation models in the \acrshort{ml} community. These models could be effectively applied  \citep{zhu2024foundations}, typically in a label-efficient manner, when specifically fine-tuned for methane detection.

\section*{Acknowledgments}

\subsection*{Author Contributions} 
Conceptualization – ET. Methods – ET, SZ (traditional approaches). Writing (original draft) – ET. Writing (review and editing) – all authors. Supervision – AK, KH, RS, XZ. Project administration – ET.

\subsection*{Funding}

This research has been enabled by OHB Digital Connect GmbH (also involved in the decision to submit the article for publication) and selected via the Open Space Innovation Platform (https://ideas.esa.int) as a Co-Sponsored Research Agreement and carried out under the Discovery programme of, and funded by, the European Space Agency.

\subsection*{Conflicts of Interest}
The authors declare that there is no conflict of interest regarding the publication of this article. 

\bibliographystyle{elsarticle-harv} 
\bibliography{sources.bib}

\clearpage
\appendix

\section{List of Acronyms}
\label{chap:acronym_list}

\printglossary[type=\acronymtype, title=List of Acronyms, toctitle=Acronyms]

\end{document}